\documentclass{article}

\usepackage{microtype}
\usepackage{booktabs} 

\usepackage{hyperref}

\usepackage[accepted]{icml2025}

\usepackage{amsmath}
\usepackage{amssymb}
\usepackage{mathtools}
\usepackage{amsthm}
\usepackage{wrapfig}
\usepackage{subcaption}
\usepackage{multirow}
\usepackage{colortbl}
\usepackage{arydshln}
\definecolor{mylightblue}{rgb}{0.827,0.937,0.945}
\definecolor{mydarkgray}{rgb}{0.66,0.66,0.66}
\definecolor{mygray}{rgb}{0.85,0.85,0.85}
\usepackage{xspace}
\def\methodshort{PrefixQuant\xspace}

\usepackage[capitalize,noabbrev]{cleveref}

\theoremstyle{plain}

\theoremstyle{definition}

\theoremstyle{remark}

\usepackage[pdftex]{graphicx}
\usepackage[textsize=tiny]{todonotes}

\begin{document}

\twocolumn[
\icmltitle{PrefixQuant: Eliminating Outliers by Prefixed Tokens for Large Language Models Quantization}





\begin{icmlauthorlist}
\icmlauthor{Mengzhao Chen}{hku}
\icmlauthor{Yi Liu}{}
\icmlauthor{Jiahao Wang}{hku}
\icmlauthor{Yi Bin}{tju}
\icmlauthor{Wenqi Shao}{shlab}
\icmlauthor{Ping Luo}{hku}
\end{icmlauthorlist}

\icmlaffiliation{hku}{The University of Hong Kong}
\icmlaffiliation{shlab}{Shanghai AI Laboratory}
\icmlaffiliation{tju}{Tongji University}

\icmlcorrespondingauthor{Wenqi Shao}{shaowenqi@pjlab.org.cn}
\icmlcorrespondingauthor{Ping Luo}{pluo@cs.hku.hk}

\icmlkeywords{Machine Learning, ICML}

\vskip 0.3in
]



\printAffiliationsAndNotice{}  

\begin{abstract}
Existing weight-activation quantization methods for Large Language Models (LLMs) primarily address channel-wise outliers but often neglect token-wise outliers, which limits the accuracy of quantized models.
In this work, we propose PrefixQuant, a novel quantization method that achieves state-of-the-art performance across various precision levels (W4A4KV4 and W4A8KV4) and granularities (dynamic and static quantization) by effectively isolating token-wise outliers. First, PrefixQuant eliminates token-wise outliers by prefixing outlier tokens in the KV cache, a process that is training-free and highly efficient (\emph{e.g.}, 1 minutes for Llama-3-70B). Second, PrefixQuant introduces new trainable parameters for block-wise training to compensate for quantization error.
Our experiments show that PrefixQuant significantly outperforms existing dynamic quantization methods, even under coarser static quantization settings. For instance, PrefixQuant achieves an average accuracy improvement of $+3.08$ and $+2.85$ points over SpinQuant (dynamic quantization) on five zero-shot reasoning tasks under dynamic and static quantization settings, respectively, on W4A4KV4 Llama-3-8B. 
Additionally, we demonstrate up to $2.74\times$ prefilling speedup and $2.16\times$ decoding speedup for LLMs using W4A4 PrefixQuant.
Our code is available at \url{https://github.com/ChenMnZ/PrefixQuant}.

\end{abstract}

\section{Introduction}
\begin{figure}[!ht]
    \centering
    \includegraphics[width=0.5\linewidth]{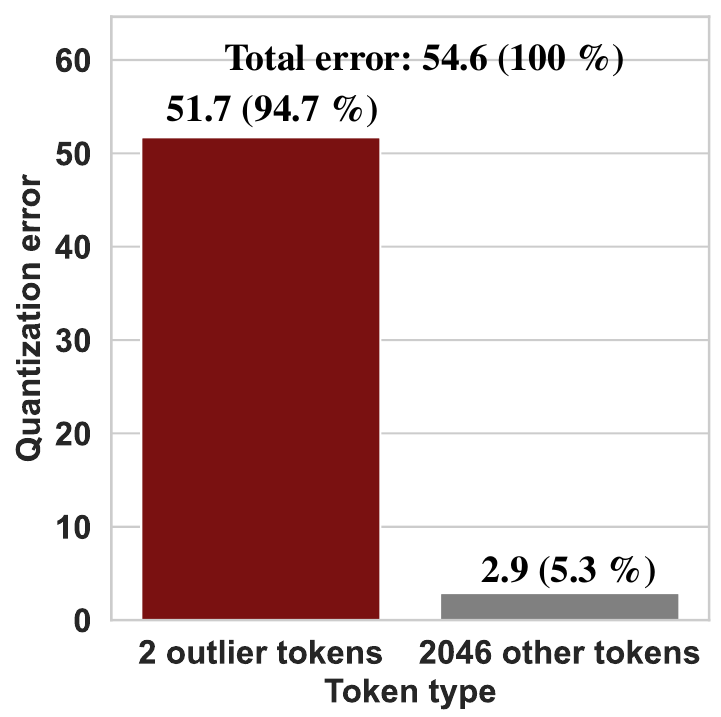}
    \caption{\textbf{4-bit 
 per-token dynamic quantization error in 2048 input context length.} Two outlier tokens account for 94.7\% of quantization error
 , while the remaining 2046 tokens contribute only 5.4\%. Quantization error is measured in the output of Llama-2-7B 2-nd transformer block   through mean square error (MSE). }
    \label{fig:quantization-loss}
\end{figure}
Recently, Large Language Models (LLMs)\citep{llama,gpt4} demonstrate remarkable capabilities across various tasks. 
However, their large parameters and computational demands pose significant challenges for deployment. This makes quantization~\citep{gptq,awq,omniquant} a crucial technology for reducing memory usage and speeding up inference~\citep{llm-unveil}.
\begin{figure*}[!ht]
    \centering
    \includegraphics[width=0.8\linewidth]{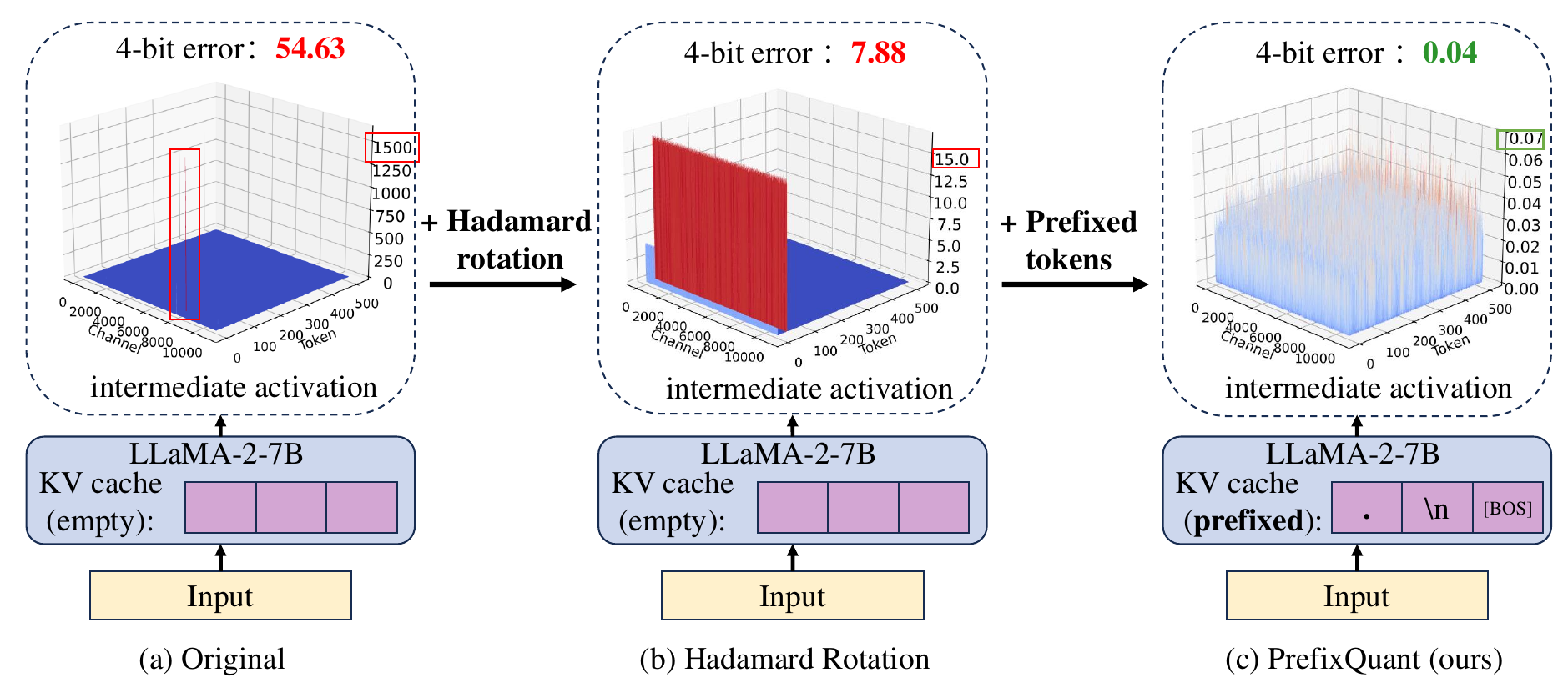}
    \caption{\textbf{Comparison of proposed \methodshort with existing methods.} This figure shows the intermediate input activation of the 2-nd down\_proj linear layer in Llama-2-7B using different methods. Quantization error is measured in the output of Llama-2-7B 2-nd transformer block through mean square error with 4-bit per-token dynamic quantization.
    The original distribution has significant outliers larger than 1,500 (left), leading 54.63 quantization error. The previous method with Hadamard rotation~\citep{quarot} reduces outliers to nearly 15 (middle) but still suffers from 7.88 quantization error. We propose \methodshort (right), which prefixes some specific tokens in KV cache to isolate outliers, reducing the maximum to nearly 0.07, significantly improving quantization error to 0.04. }
    \label{fig:teaser}
    \vspace{-0.5cm}
\end{figure*}

Despite advancements, outliers in LLMs activations can lead to significant quantization errors and accuracy degeneration. Many current methods address this by focusing on alleviating channel-wise outliers \citep{llm-int8} through techniques like channel-wise scaling \citep{smoothquant,omniquant,os+}, mixed-precision quantization \citep{llm-int8,atom}, Hadamard rotation \citep{quarot,spinquant}, or channel-level assembly \citep{qllm}. However, activations of LLMs include not only channel-wise outlier but also some massive activation~\cite{massive} only occur in a few tokens, and can be termed as token-wise outliers.
For example, Figure~\ref{fig:quantization-loss} shows that 2 outlier tokens among 2048 tokens contribute 94.7\% of the quantization error. Figure~\ref{fig:teaser}(a) provides a more detailed analysis, revealing that a few tokens have extreme values exceeding 1,000, resulting in a quantization error of $54.63$.
The current state-of-the-art method, Hadamard rotation\citep{quarot}, redistributes outlier values across all channels, reducing the maximum value of outlier tokens from over 1,000 to approximately 15 (see Figure~\ref{fig:teaser}~(b)). However, the magnitude of outlier tokens remains hundreds of times larger than normal tokens, leading to a quantization error of $7.88$.

In this paper, we propose \methodshort, an efficient method to isolate token-wise outliers for more accurate quantization. \methodshort is based on a key observation: \textbf{Prefixing high-frequency outlier tokens at the beginning of the input sequence constrains token-wise outliers to only occur in the prefixed tokens.}
Since prefixed tokens remain consistent across all inputs, \methodshort performs offline prefilling of these tokens and stores their KV cache. This stored KV cache can then be reused for all inputs, effectively avoiding token-wise outliers during the forward pass.
Furthermore, the detection of prefixed tokens is efficient and does not require any retraining, unlike prior methods~\citep{massive,remove-outliers}. For example, this process completes in just 12 seconds for Llama-2-7B.
As shown in Figure~\ref{fig:teaser}(c), \methodshort effectively eliminates outliers and reduces the quantization error from 7.88 (using QuaRot) to 0.04.
Additionally, we introduce a block-wise fine-tuning \citep{omniquant,efficientqat} to compensate for quantization error by jointly training weights and quantization parameters. For static activation quantization, the quantization parameters are inherently trainable. However, dynamic activation quantization lacks trainable quantization parameters. To address this, we propose learnable activation clipping to enable training for dynamic activation quantization.

Since PrefixQuant is compatible with various quantization schemes, we introduce two settings for PrefixQuant (see Table~\ref{tab:quantization_settngs}): O1 for dynamic quantization and O2 for static quantization.
Experiments show that \methodshort significantly outperforms existing methods~\citep{quarot,smoothquant,qserve} under the same dynamic quantization setting. Furthermore, PrefixQuant even surpasses prior dynamic quantization methods while utilizing the more efficient static quantization.
For instance, \methodshort achieves an average accuracy improvement of $+3.08$ and $+2.85$ points over SpinQuant (dynamic quantization) on five zero-shot reasoning tasks under dynamic and static quantization settings, respectively, on W4A4KV4 Llama-3-8B.
To the best of our knowledge, \methodshort is the first method to surpass prior per-token dynamic quantization methods~\citep{quarot,smoothquant,qserve} using the coarser per-tensor static quantization.
We also benchmark the end-to-end inference of W4A4 quantization, where \methodshort achieves a $2.74\times$ prefilling speedup and a $2.16\times$ decoding speedup compared to FP16 models.
We hope \methodshort inspires future developments in LLM compression.

\begin{table*}[!ht]
    \centering
    \caption{Quantization setting of the baselines and PrefixQuant. All group-wise quantization set group size as 128. For PrefixQuant, O1 is the same as existing methods for fair comparisons, and O2 is more efficient than O1 (\emph{i.e.} lower latency).}
    \label{tab:quantization_settngs}
    \begin{tabular}{cccc}
    \toprule
    \bf{Method} & \bf{Weight} & \bf{Activation} & \bf{KV Cache} \\
    \midrule
    SmoothQuant & per-channel  & per-token dynamic &  per-token dynamic \\
    Atom &  group-wise  & group-wise dynamic & group-wise dynamic \\
    QoQ;QuaRot;SpinQuant;SpinQuant & per-channel  & per-token dynamic & group-wise dynamic\\
    \hline
    \textbf{\methodshort-O1} & per-channel & per-token dynamic & group-wise dynamic\\
    \textbf{\methodshort-O2} & per-channel & per-tensor static & per-head  static\\
    \bottomrule
    \end{tabular}
\end{table*}

\section{Related Works}
This section discusses works related to enhancing quantization accuracy by eliminating activation outliers.

\textbf{Channel-Wise Outliers.}  Activation outliers often recur in the same channels across tokens. \cite{llm-int8} addresses this by isolating outlier channels with 16-bit precision, while Atom~\citep{atom} and QUIK~\citep{quik} adopt similar mixed-precision strategies. Other methods, like SmoothQuant~\citep{smoothquant}, OmniQuant~\citep{omniquant}, and Outlier Suppression~\citep{outlier,outlier-plus}, scale activations to weights on a channel-wise basis. QLLM~\citep{qllm} splits outlier channels into sub-channels, and QuaRot~\citep{quarot} redistributes outliers using random Hadamard rotation, later improved by SpinQuant~\citep{spinquant}, which trains the orthogonal matrix. 
In contrast, our work focuses on token-wise outliers and is orthogonal to these channel-wise methods.

%

\textbf{Token-Wise Outliers.} 
The SoftMax function in self-attention prevents zero attention scores, causing unnecessary scores for special tokens and leading to token-wise outliers~\citep{massive,streamingllm,gu2024attention}. StreamingLLM~\citep{streamingllm} and LM-infinite~\citep{lm-infinite} retain initial tokens for long-context generation, while our \methodshort isolates outliers by carefully selecting prefixed tokens in the KV-cache for quantization.
Unlike training-based methods~\citep{remove-outliers,massive} that modify SoftMax behavior or add attention bias, our \methodshort isolates outliers without retraining.
Closest works, like QFeP~\citep{qfep} and CushionCache~\citep{cushioncache}, rely on costly grid searches (e.g., 12 hours for Llama-3-8B), while \methodshort completes this in 12 seconds. Furthermore, unlike prior works~\citep{massive,cushioncache} focusing on large-value outliers, \methodshort also identifies extremely small-value outliers in self-attention queries and keys.


\section{Preliminaries}

Quantization in LLMs involves weight, activation, and KV cache quantization. Weight quantization~\citep{efficientqat} and KV cache quantization~\citep{kivi} reduce memory usage and speed up memory-bound computations~\citep{llm-unveil}. Combining weight and activation quantization enables low-bit matrix manipulation to accelerate computation-bound tasks~\citep{llm-unveil}. Specifically, the quantization process is:
\begin{align}\label{eq:quant}
    \mathbf{X}_{\texttt{INT}} &= \mathrm{clamp}\left(\lfloor \frac{\mathbf{X}}{s} \rceil + z, 0, 2^{N}-1\right), \\
    \text{where} \quad s &= \frac{\gamma\text{max}(\mathbf{X})-\beta\text{min}(\mathbf{X})}{2^{N}-1}, z = -\lfloor \frac{\beta\textbf{min}(\mathbf{X})}{s} \rfloor
\end{align}
where $\lfloor \cdot \rceil$ denotes rounding operation, $N$ is the target bit number, $\mathbf{X}_{\texttt{INT}}$ and $\mathbf{X}$ are the quantized integer and full-precision tensor, respectively. $\mathbf{s}$ and $\mathbf{z}$ are quantization parameters, for the step size and zero values, respectively. $\gamma \in [0,1]$ and $\beta \in [0,1]$ are clipping factors. 

\textbf{Dynamic and Static.} 
Activation quantization is divided into dynamic and static quantization based on how quantization parameters are calculated. Specifically, dynamic quantization calculates $s$ and $z$ online during inference, offering better adaptability to different distributions. In contrast, static quantization precomputes $s$ and $z$ offline through calibration datasets, leading to more efficient inference and more feasible operator fusion~\citep{whitepaper}. 

\textbf{Initialization of Quantization Parameters.} The classical approach uses max–min initialization, where both $\gamma$ and $\beta$ are set to $1$. To better balance clipping error and rounding error~\citep{awq,qserve}, we initialize $\gamma$ and $\beta$ using MSE-based grid search for both weight and activation quantization in our experiments. Specifically, for per-token dynamic quantization, $\gamma$ and $\beta$ are shared across all tokens within the same layer. For per-tensor static quantization, we directly perform a grid search for the quantization parameters $s$ and $z$ instead of optimizing the clipping factors.

\textbf{Hadamard Rotation.} 
Random Hadamard rotation~\citep{quarot,spinquant} addresses channel-wise outliers. Our method focus on removing token-wise outliers. Therefore, We build our method upon the Hadamard rotation technique, and the detailed is provided in Sec.~\ref{sec:details_of_rotation}.

\begin{figure*}[!ht]
    \centering
\includegraphics[width=0.8\linewidth]{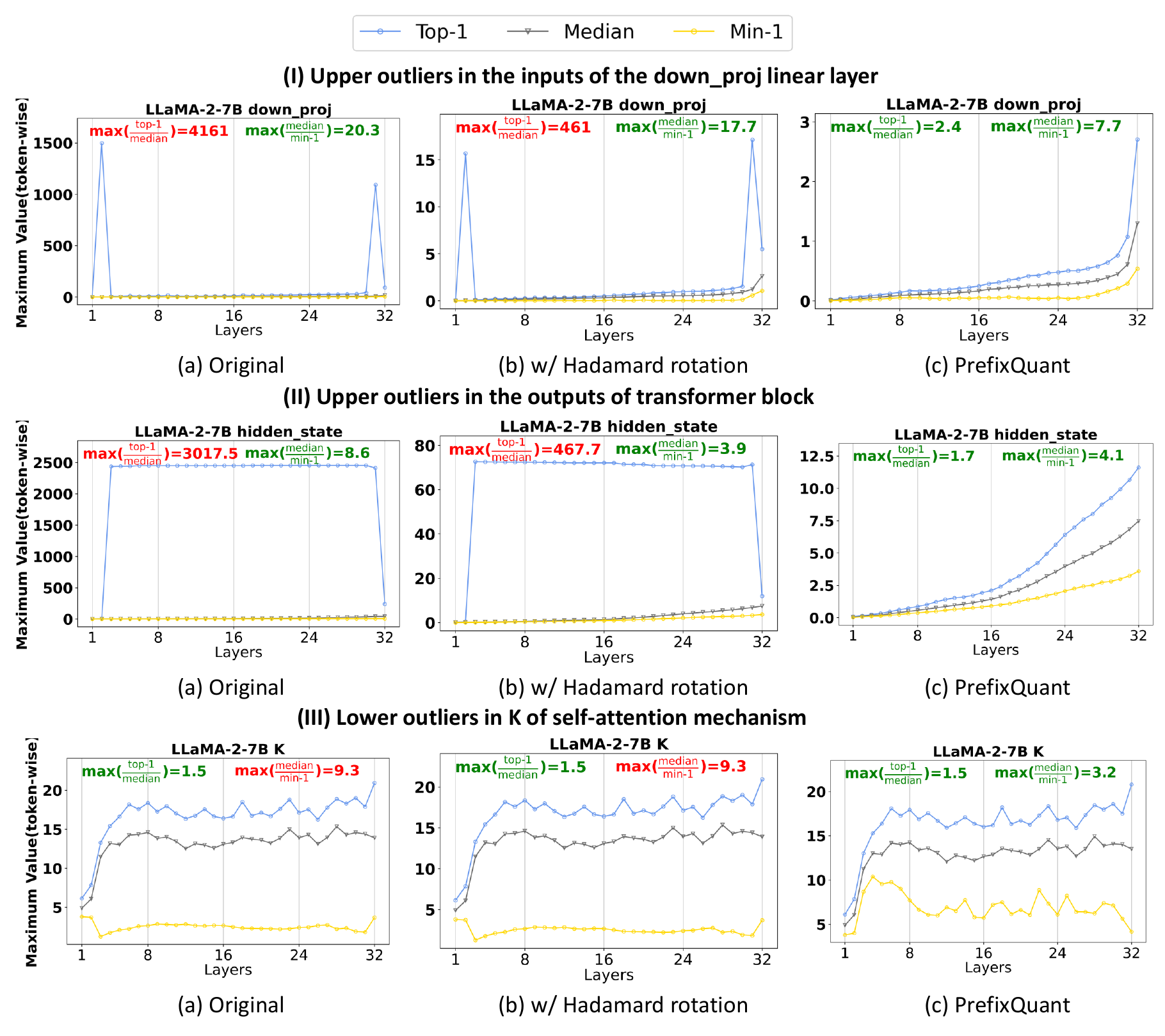}
    \caption{
    Example of token-wise outliers. We present (I)(II) upper outliers and (III) lower outliers. Top-1, Medium, Min-1 indicate the largest, median, and smallest values among token-wise maximum values, respectively. We also calculate the the ratios of $\frac{\text{Top-1}}{\text{Median}}$ and $\frac{\text{Median}}{\text{Min-1}}$ in each layer, and report the maximum ratio across all layers .  A lower ratio indicates a more uniform distribution. we take Llama-2-7B as an example here, more visualizatiosn about other models can be find in Sec.\ref{sec:more_visualization}.}
    \label{fig:upper-lowwer-outliers}
\end{figure*}

\section{\methodshort}
In this section, we present the proposed PrefixQuant methods. Sec.\ref{sec:outlier-token} describes the characteristics of outlier tokens in LLMs. Sec.\ref{sec:prefixed_outliers} proposes a solution to isolate these outlier tokens, creating a more quantization-friendly distribution. Finally, Sec.\ref{sec:fine-tuning} introduces block-wise optimization to further reduce quantization error.

\subsection{Deep Exploration of Outlier Tokens}\label{sec:outlier-token}

Both channel-wise and token-wise outliers can cause significant quantization error. While channel-wise outliers have been thoroughly explored and addressed in prior research~\citep{quarot}, this discussion focuses on token-wise outliers.

\textbf{Definition of Outlier Token.} Let $\mathbf{X} \in \mathbb{R}^{T \times C}$ represent the absolute values of a token sequence, where $T$ is the number of tokens and $C$ is the dimension size. We compute the token-wise maximum values $\mathbf{M} \in \mathbb{R}^{T}$, where each element $\mathbf{M}_{i}$ indicates the maximum value of the $i$-th token. The outlier degree of a token is then measured by comparing its maximum value to the median of $\mathbf{M}$:

\begin{equation}\label{eq:definition}
R_i = \frac{\mathbf{M}_{i}}{\text{median}(\mathbf{M})}.
\end{equation}

Then, the $i$-th token is classified as an outlier if $R_i$ deviates significantly from 1 (\emph{i.e.}, either much larger or much smaller than 1). Specifically, we define an \textit{upper outlier token} when $R_i > \eta_{1}$ and a \textit{lower outlier token} when $R_i^{-1} > \eta_{2}$. In our experiments, we set $\eta_{1} = 64$ and $\eta_{2} = 8$.

\textbf{Visualization of Outlier Tokens.} To better illustrate the outlier degree, we further define $\max\left(\frac{\text{top-1}}{\text{median}}\right)$ and $\max\left(\frac{\text{median}}{\text{min-1}}\right)$ as the maximum $R_i$ and maximum $R_i^{-1}$ across different layers, respectively.
Following the definition of $R_i$ in Eq.~(\ref{eq:definition}), a larger $\max\left(\frac{\text{top-1}}{\text{median}}\right)$ indicates the presence of extreme upper outliers, while a larger $\max\left(\frac{\text{median}}{\text{min-1}}\right)$ reflects the presence of extreme lower outliers.
Specifically, we identify the following outlier tokens:

\textbf{1) Upper outlier tokens in inputs of down\_proj layers and outputs of transformer blocks.}
As shown in Figure~\ref{fig:upper-lowwer-outliers}(I.a), the input activations of the down\_proj layers exhibit significant upper outliers, with $\max\left(\frac{\text{top-1}}{\text{median}}\right) = 4161$. Although Hadamard rotation (Figure~\ref{fig:upper-lowwer-outliers}(I.b)) reduces this ratio to 461, it still indicates a large gap compared to normal tokens. A similar phenomenon is observed in the outputs of transformer blocks, as shown in Figure~\ref{fig:upper-lowwer-outliers}(II). These outliers not only lead to larger quantization errors but also cause instability during block-wise fine-tuning.

\textbf{2) Lower outlier tokens in $\mathbf{Q}$/$\mathbf{K}$.}
In Figure~\ref{fig:upper-lowwer-outliers}(II.a), $\mathbf{K}$ displays a distinct outlier pattern compared to the inputs of linear layers. Instead of having large magnitudes, some tokens exhibit extremely small values. Specifically, $\mathbf{K}$ has $\max\left(\frac{\text{top-1}}{\text{median}}\right) \approx 1.5$, but $\max\left(\frac{\text{median}}{\text{min-1}}\right) > 9$. Furthermore, as shown in Figure~\ref{fig:upper-lowwer-outliers}(II.b), Hadamard rotation does not mitigate these lower outliers. Similar lower outliers are also observed in $\mathbf{Q}$, as shown in Figure~\ref{fig:llama-2-7b-output-all}.

Additionally, we observe that both upper outlier tokens and lower outlier tokens correspond to tokens at the same position in the sequence, but they exhibit different patterns in different modules. Therefore, we focus on analyzing upper outlier tokens due to their stronger prominence and ease of detection.

\begin{figure*}[!ht]
    \centering
    \begin{subfigure}[b]{0.21\textwidth}
    \centering
        \includegraphics[width=\textwidth]{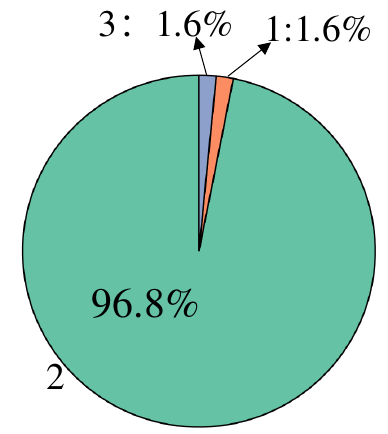}
        \vspace{-0.7em}
        \caption{Number of outlier tokens\\~}
        \label{fig:number}
    \end{subfigure}
    \begin{subfigure}[b]{0.23\textwidth}
    \centering
        \includegraphics[width=\textwidth]{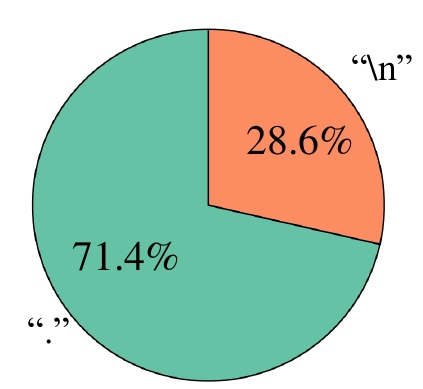}
        \vspace{-0.7em}
        \caption{Content of outlier tokens (exclude position 0)}
        \label{fig:content}
    \end{subfigure}
    \begin{subfigure}[b]{0.25\textwidth}
    \centering
        \includegraphics[width=\textwidth]{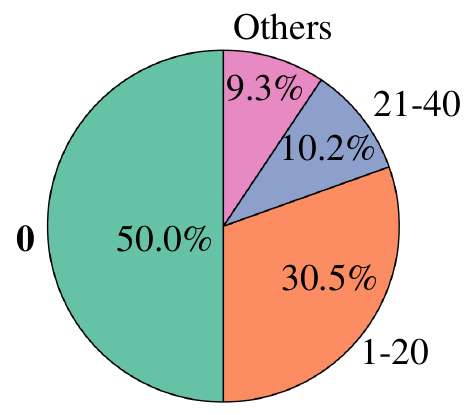}
        \vspace{-0.7em}
        \caption{Position index of outlier tokens\\~}
        \label{fig:position}
    \end{subfigure}
    \begin{subfigure}[b]{0.23\textwidth}
    \centering
        \includegraphics[width=\textwidth]{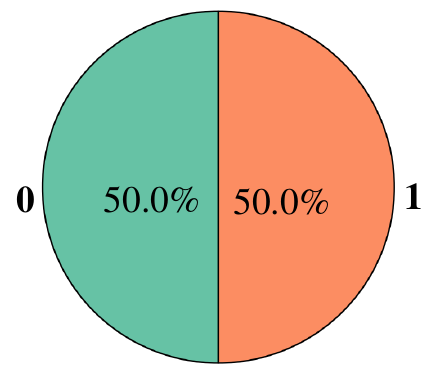}
        \vspace{-0.7em}
        \caption{Position index of outlier tokens \textbf{w/ prefixed tokens}}
        \label{fig:position-prefix}
    \end{subfigure}
    \caption{\textbf{Explorations of outlier tokens in Llama-2-7B.} (a) Outlier token only exits in nearly 2 positions in the overall input sequence. (b) Excluding token in position 0, outlier tokens only exits in `.'' or ``\textbackslash n'' tokens.  (c) Outlier tokens consistently occur in the starting token (position 0) and another front but un-predictable position index. (d) Prefixing the input sequence with high-frequency outlier tokens (``.\textbackslash n'') can constraint the outlier tokens only exit in position 0 and 1.  }
    \label{fig:llama-2-7b-deep-exploration}
\end{figure*}

\textbf{Characters of outlier tokens.} We further investigate the characteristics of these outlier tokens, including the number of outlier tokens in an input sequence, their positions, and their content (text):
\begin{itemize}
    \item \textbf{Number}:  We determine the number of outlier tokens in a small calibration dataset. Specifically, we compute the average outlier token count $\mathbf{O} \in \mathbb{R}^{b}$ for each transformer block according to compare Eq~(\ref{eq:definition}) with the outlier threshold $\eta_1$, where $b$ is the total number of transformer blocks. Since outlier tokens are nearly consistent across layers that contain them, we simply set the number of outlier tokens as $o = \lceil \max(\mathbf{O}) \rceil$. Consistent with Massive Attention~\citep{massive}, we find that outlier tokens appear in only a small fraction of positions (\emph{e.g.} 2 for Llama-2-7B) within the input sequence, as shown in Figure~\ref{fig:number}.
    
    \item \textbf{Position}: We observe that the initial tokens are outlier tokens across almost all models, aligning with findings on attention sinks~\citep{streamingllm}. Additionally, Figure~\ref{fig:position} shows that, apart from the initial tokens, some other tokens near the beginning of the sequence are also outlier tokens. Unlike outlier channels, which occur at fixed channel indices~\citep{llm-int8}, the positions of outlier tokens depend on the input sequence and vary significantly. As a result, it is not feasible to identify outlier tokens offline for mixed-precision quantization.

    \item \textbf{Content (text)}: Initial tokens are consistently outlier tokens, regardless of their content. Thus, we focus on outlier tokens that are not initial tokens to analyze their content.
Some models, such as Llama-3-8B and Qwen-2-7B, exhibit outlier tokens only at the initial positions. However, certain models display outlier tokens not only at the start of the input sequence but also in low-semantic tokens. For example, Llama-2-7B shows outlier tokens in both initial and delimiter tokens (e.g., .'' or \textbackslash n''), as illustrated in Figure~\ref{fig:content}. Notably, tokens corresponding to the same text may exhibit different patterns depending on their position in the sequence. For instance, low-semantic tokens may behave as outlier tokens at the front of the sequence but appear as normal tokens in other positions.
\end{itemize}


\subsection{Prefixed Outliers}\label{sec:prefixed_outliers}
Given that the number of outlier tokens is limited and they typically occur at the beginning of the input sequence, we propose a method to prefix high-frequency outlier tokens in the input sequence. This approach constrains outlier tokens to the prefixed tokens. Furthermore, these prefixed tokens can be directly stored in the KV cache, enabling more efficient computation, as shown in Figure~\ref{fig:teaser}.

\newcommand{\visibleSpace}{\rule{0.5em}{0.5pt}}
\begin{figure}[!h]
    \centering
    \vspace{-0pt}
    \caption{Prefixed tokens in KV cache across different models. [BOS] indicates the special token for beginning of sequence(\emph{e.g.} ``$<$s$>$'' for Llama-2 and ``$|$begin\_of\_text$|$`` for Llama-3). Note that the following ``\visibleSpace'' represents space.}
    \begin{tabular}{lcr}
    \toprule
    \multirow{2}{*}{\bf{Model}} & \multicolumn{2}{c}{\bf{Prefixed token}}   \\
    \cmidrule(lr){2-3}
     &   \bf{Number} & \bf{Content} \\
     \midrule
     Llama-2-7B &  3 & .\textbackslash n[BOS] \\
     Llama-2-13B & 3 & the.[BOS]\\
     Llama-2-70B & 4 & \textbackslash n’'[BOS]\\
     Llama-3-8B(-Instruct) & 1 & [BOS] \\
     Llama-3-70B(-Instruct) & 3 & ,\visibleSpace[BOS] \\
     Mistral-v0.3-7B & 4 & \textbackslash n.to[BOS]\\
     Qwen-2-7B & 1 & [BOS] \\
     \bottomrule
    \end{tabular}
    \label{tab:prefixed_token}
\end{figure}

\textbf{What token is added as a prefix.}  
To determine which tokens to add as a prefix, we firstly analyze the number of outlier tokens $o$. We find that prefixing the top-$o$ high-frequency\footnote{The frequencies are calculated excluding the initial token.} outlier tokens successfully constrains outliers to the prefixed tokens, as illustrated in Figure~\ref{fig:position-prefix}. For special cases, such as models like Llama-3-8B and Qwen-2-7B where outlier tokens only appear as initial tokens, we set the prefix token to ''[BOS]''. Additionally, for consistency, we also include ''[BOS]'' as the last prefixed token for all models.
The detailed prefixed tokens used for different models are listed in Table~\ref{tab:prefixed_token}.

\textbf{Computation of prefixed tokens in KV cache.} 
In the auto-regressive inference pipeline of LLMs, we directly store these prefixed tokens in the KV cache to prevent new outlier tokens from being generated during inference. Specifically, given the input query, key, and value matrices $\mathbf{Q}, \mathbf{K}, \mathbf{V} \in \mathbb{R}^{T \times C}$, the self-attention mechanism with prefixed tokens in the KV cache is formulated as:
\begin{equation}\label{eq-attn-bias}
    \text{Attention}(\mathbf{Q}, \mathbf{K}, \mathbf{V};\, \mathbf{k'}, \mathbf{v'}) = \text{Softmax}\left(\frac{\mathbf{Q} \begin{bmatrix}
        \mathbf{K}^{T}\,\,\,\mathbf{k'}
    \end{bmatrix}}{\sqrt{d}}\right)\begin{bmatrix}
        \mathbf{V} \\ 
        \mathbf{v'}^{T}   \end{bmatrix}
\end{equation}
Here, $\mathbf{k'}, \mathbf{v'} \in \mathbb{R}^{o \times C}$ are the prefixed tokens stored in the KV cache. We compute $\mathbf{k'}$ and $\mathbf{v'}$ during a one-time prefilling process using the full-precision model. These prefixed tokens are then stored in the KV cache and reused during inference by quantized models. Notably, the prefixed tokens in the KV cache remain in full precision, even when used with quantized models.

\textbf{Distribution changing after setting prefixed tokens.} 
As shown in Figure~\ref{fig:upper-lowwer-outliers}, prefixing outlier tokens in the KV cache significantly improves the distribution. Specifically, the $\max(\frac{\text{top-1}}{\text{median}})$ ratio of the down\_proj inputs decreases from 461 to 2.4 and the $\max(\frac{\text{median}}{\text{min-1}})$ ratio of $\mathbf{Q}$/$\mathbf{K}$ decreases from $>9$ to $<3.5$.

\subsection{Block-wise Fine-tuning}\label{sec:fine-tuning}
Recent studies show that block-wise fine-tuning~\citep{omniquant,efficientqat} improves performance by accounting for inter-layer interactions~\citep{brecq}. To further enhance the performance of quantized models, we fine-tune each transformer block sequentially using a mean squared error (MSE) loss. Specifically, we introduce trainable parameters for the activation quantizer to balance the rounding and clipping errors in quantization.
For dynamic activation quantization in PrefixQuant-O1, we set the tensor-wise clipping factors as trainable. Note that the clipping factors cannot be token-wise, as long-context scenarios introduce excessive storage overhead with token-wise clipping factors. For static activation quantization in PrefixQuant-O2, the quantization parameters (scaling factors and zero-points) are inherently trainable.
For weight quantization, we follow the approach of EfficientQAT~\citep{efficientqat}, enabling the training of all weights and weight quantization parameters.

\section{Experiments}

\subsection{Setups}~\label{sec:setups}

\textbf{Baseline.} \methodshort is a versatile method applicable to any precision. We conduct experiments on two mainstream precisions: W4A8KV4, and W4A4KV4. The detailed quantization settings are illustrated in Table~\ref{tab:quantization_settngs}. PrefixQuant-O1 is consistent with existing methods for fair comparisons, and PrefixQuant-O2 targets to push the limitation of more efficient static quantization. 
We compare \methodshort with QuaRot~\citep{quarot}, Atom~\citep{atom}, DuQuant~\citep{duquant}, QoQ~\citep{qserve},  and SpinQuant~\citep{spinquant}. Following QoQ, we reproduce all these methods except SpinQuant
with Pile~\citep{pile} calibration dataset to avoid over-fitting for fair comparisons. The detailed quantization configuration and results sources of these comparison methods can be found at Sec.~\ref{sec:setting_and_results_source}. 

\begin{table}[!t]
\renewcommand{\arraystretch}{0.2}
\centering
\caption{\textbf{W4A4KV4 results on Llama models.} ``PPL'' indicates WikiText2 perplexity measured with context length 2048. ``Acc.'' indicates the average zero-shot accuracy on 5 common-sense reasoning tasks. Grayed results use Wikitext2 as calibaration dataset.}
\label{tab:w4a4}
\centering
    \begin{tabular}{cccc}
    \toprule
    \bf{Model} & \bf{Method} & \bf{PPL} & \bf{Acc.} \\
    \midrule
    \multirow{7}{*}{2-7B} & FP16& 5.47 & 69.04 \\
    \cdashline{2-4} 
    & Atom & 6.12 & 59.73 \\
    & QuaRot & 6.19 & 64.69\\
    & DuQuant & 6.20 & 66.25 \\
    & SpinQuant &  \textcolor{gray}{5.95} &  \textcolor{gray}{65.35} \\
    & \cellcolor{mylightblue} \methodshort-O1  &\cellcolor{mylightblue} \textbf{5.93} &  \cellcolor{mylightblue} \textbf{66.74} \\
    & \cellcolor{mylightblue} \methodshort-O2 & \cellcolor{mylightblue} 6.01 &  \cellcolor{mylightblue} 66.37 \\
    \midrule
    \multirow{7}{*}{2-13B} & FP16& 4.88  & 71.73 \\
    \cdashline{2-4} 
    & Atom & 5.31  & 63.51 \\
    & QuaRot & 5.45  & 69.01 \\
    & DuQuant & 5.39 & 69.13 \\
    & SpinQuant &  \textcolor{gray}{5.24} &  \textcolor{gray}{69.24} \\
    & \cellcolor{mylightblue} \methodshort-O1  &\cellcolor{mylightblue} \textbf{5.24} &  \cellcolor{mylightblue} 70.05 \\
    & \cellcolor{mylightblue} \methodshort-O2 & \cellcolor{mylightblue} 5.32 & \cellcolor{mylightblue} \textbf{70.36} \\
    \midrule
    \multirow{7}{*}{2-70B} & FP16& 3.32 & 76.72 \\
    \cdashline{2-4} 
    & Atom & 3.73 & 67.52\\
    & QuaRot & 3.83 & 75.43 \\
    & DuQuant & 3.77 & 74.75 \\
    & SpinQuant &  \textcolor{gray}{3.70} &  \textcolor{gray}{75.19} \\
    & \cellcolor{mylightblue} \methodshort-O1  &\cellcolor{mylightblue} \textbf{3.62} &  \cellcolor{mylightblue} \textbf{76.23} \\
    & \cellcolor{mylightblue} \methodshort-O2 & \cellcolor{mylightblue} 3.81 &  \cellcolor{mylightblue} 75.48 \\
    \midrule
    \multirow{7}{*}{3-8B} & FP16& 6.14 & 72.71 \\
    \cdashline{2-4} 
    & Atom & 7.76\\
    & QuaRot & 8.41 & 65.15 \\
    & DuQuant & 8.14 & 67.13 \\
    & SpinQuant &  \textcolor{gray}{7.36} &  \textcolor{gray}{68.23} \\
    & \cellcolor{mylightblue} \methodshort-O1  &\cellcolor{mylightblue} \textbf{7.26} &  \cellcolor{mylightblue} \textbf{71.31} \\
    & \cellcolor{mylightblue} \methodshort-O2 & \cellcolor{mylightblue} 7.43 &  \cellcolor{mylightblue} 71.08 \\
    \midrule
    \multirow{5}{*}{3-70B} & FP16& 2.85  & 80.03 \\
    \cdashline{2-4} 
    & QuaRot & 6.82 & 68.39 \\
    & DuQuant & 5.67 & 74.89 \\
    & \cellcolor{mylightblue} \methodshort-O1  &\cellcolor{mylightblue} \textbf{4.16} &  \cellcolor{mylightblue} 77.08 \\
    & \cellcolor{mylightblue} \methodshort-O2 & \cellcolor{mylightblue} 4.41 &  \cellcolor{mylightblue} \textbf{77.18} \\
    \bottomrule
    \end{tabular}
\end{table}
\begin{table}[!t]
\renewcommand{\arraystretch}{0.2}
\centering
\caption{\textbf{W4A8KV4 results on Llama models.} Refer Table~\ref{tab:w4a4} for the metric setting.}
\label{tab:w4a8}
\begin{tabular}{cccc}
\toprule
\bf{Model} & \bf{Method} & \bf{PPL} &  \bf{Acc.} \\
\midrule
\multirow{5}{*}{2-7B} & FP16& 5.47 & 69.04 \\
\cdashline{2-4} 
& QoQ & 5.75 &  67.22 \\
& QuaRot & 5.73  & 67.11 \\
& \cellcolor{mylightblue} \methodshort-O1 & \cellcolor{mylightblue} \textbf{5.67}  & \cellcolor{mylightblue} 68.04 \\
& \cellcolor{mylightblue} \methodshort-O2 & \cellcolor{mylightblue} 5.68  & \cellcolor{mylightblue} \textbf{68.09} \\
\midrule
\multirow{5}{*}{2-13B} & FP16& 4.88  & 71.73 \\
\cdashline{2-4} 
& QoQ & 5.12 & 70.56 \\
& QuaRot & 5.07  & 69.96 \\
& \cellcolor{mylightblue} \methodshort-O1 & \cellcolor{mylightblue} \textbf{5.05}  & \cellcolor{mylightblue} \textbf{71.25} \\
& \cellcolor{mylightblue} \methodshort-O2 & \cellcolor{mylightblue} 5.07  & \cellcolor{mylightblue} \textbf{71.25} \\
\midrule
\multirow{5}{*}{2-70B} & FP16& 3.32 & 76.72 \\
\cdashline{2-4} 
& QoQ & 3.52  & 75.91 \\
& QuaRot & 3.46  & 76.31 \\
& \cellcolor{mylightblue} \methodshort-O1 & \cellcolor{mylightblue} \textbf{3.44}  & \cellcolor{mylightblue} \textbf{76.82} \\
& \cellcolor{mylightblue} \methodshort-O2 & \cellcolor{mylightblue} 3.50 & \cellcolor{mylightblue} 76.50 \\
\midrule
\multirow{5}{*}{3-8B} & FP16& 6.14 & 72.71 \\
\cdashline{2-4} 
& QoQ & 6.89  & 71.35 \\
& QuaRot & 6.80  & 71.68 \\
& \cellcolor{mylightblue} \methodshort-O1 & \cellcolor{mylightblue} \textbf{6.59}  & \cellcolor{mylightblue} \textbf{72.57} \\
& \cellcolor{mylightblue} \methodshort-O2 & \cellcolor{mylightblue} 6.62  & \cellcolor{mylightblue} 72.46 \\
\midrule
\multirow{5}{*}{3-70B} & FP16& 2.85  & 80.03 \\
\cdashline{2-4} 
& QoQ & 4.36 & 78.12 \\
& QuaRot & 3.73 & 78.52 \\
& \cellcolor{mylightblue} \methodshort-O1 & \cellcolor{mylightblue} \textbf{3.37}  & \cellcolor{mylightblue} 78.50 \\
& \cellcolor{mylightblue} \methodshort-O2 & \cellcolor{mylightblue} 3.43 & \cellcolor{mylightblue} \textbf{78.70} \\
\bottomrule
\end{tabular}
\end{table}
\begin{table}[!t]
\renewcommand{\arraystretch}{0.2}
    \centering
    \caption{MMLU average accuracy (zero-shot) on Llama-3-8B.}
    \begin{tabular}{ccc}
    \toprule
    Method & Precision & MMLU Acc. \\
    \midrule
    - & FP16 & 62.07 \\
    \midrule
    QuaRot & W4A4KV4 & 34.25 \\
    DuQuant & W4A4KV4 & 50.77 \\
    SpinQuant & W4A4KV4 & 51.93 \\
    \cellcolor{mylightblue}PrefixQuant-O1 & \cellcolor{mylightblue}W4A4KV4 & \cellcolor{mylightblue}\textbf{56.00} \\
    \cellcolor{mylightblue}PrefixQuant-O2 & \cellcolor{mylightblue}W4A4KV4 & \cellcolor{mylightblue}54.65 \\
    \midrule
    QuaRot & W4A8KV4 &  38.37 \\
    DuQuant & W4A8KV4 &  58.01 \\
    SpinQuant & W4A8KV4 &  58.25 \\
    \cellcolor{mylightblue}PrefixQuant-O1 & \cellcolor{mylightblue}W4A8KV4 &  \cellcolor{mylightblue}\textbf{60.49} \\
    \cellcolor{mylightblue}PrefixQuant-O2 & \cellcolor{mylightblue}W4A8KV4 &  \cellcolor{mylightblue}59.20 \\
    \bottomrule
    \end{tabular}
    \label{tab:mmlu_results}
\end{table}

\textbf{Models and datasets.} We evaluate \methodshort on the Llama-2, Llama-3, Llama-3-Instruct families, Mistral-7B-v0.3, and Qwen-2-7B models. Following previous literature~\citep{omniquant,qserve}, we assess \methodshort quantized models on language modeling and zero-shot tasks. Specifically, we evaluate on WikiText2~\citep{wikitext2} with a 2048 context length for perplexity, and on 5 zero-shot reasoning tasks, including PIQA~\citep{piqa}, ARC~\citep{arc}, HellaSwag~\citep{hellaswag}, and WinoGrande~\citep{winogrande}. We also test models on more challenge zero-shot MMLU~\citep{mmlu}. All accuracy are measured through \texttt{lm\_eval v0.4.2}~\citep{eval-harness}. For accuracy, we report \texttt{acc} for WinoGrande and \texttt{acc\_norm} for HellaSwag, Arc\_Challenge, Arc\_Easy, and PIQA, following Qserve~\citep{qserve}.

\begin{table*}[!ht]
\renewcommand{\arraystretch}{0.3}
    \centering
    \caption{Ablation study on quantization techniques used in PrefixQuant. The model used here is Llama-3-8B, and WikiText2 perplexity with 2048 context length is reported. Both PrefixQuant-O1  and PrefixQuant-O2 are start from the ``Base''.}
    \begin{tabular}{lcccc}
    \toprule
    & \bf{Method} & \bf{Activation Quant.} & \bf{W4A8KV4} &  \bf{W4A4KV4} \\
    \midrule
    & QuaRot & dynamic & 6.75 & 8.33 \\
    \midrule
    \multirow{3}{*}{Base} & RTN & dynamic & 12.66 & 1282.34 \\
    & + rotation & dynamic & 10.88 & 24.98 \\
    & + Grid search initialization & dynamic &  9.08 & 11.70 \\
    \cdashline{1-5}
    \multirow{2}{*}{PrefixQuant-O1}& + \textbf{prefixed outliers} & \textbf{dynamic} & \textbf{6.81} & \textbf{7.53} \\
    & + \textbf{block-wise fine-tuning} & \textbf{dynamic} &  \textbf{6.59} & \textbf{7.23} \\
    \cdashline{1-5}
    \multirow{3}{*}{PrefixQuant-O2} & + static quantization & static & 29.07 & 141.02 \\
    & + \textbf{prefixed outliers} & \textbf{static} & \textbf{6.90} & \textbf{7.93} \\
    & + \textbf{block-wise fine-tuning} & \textbf{static} & \textbf{6.62} & \textbf{7.41} \\
    \bottomrule
    \end{tabular}
    \vspace{-0.3em}
    \label{tab:ablation_techniques}
\end{table*}

\textbf{Grid Search Initialization Setting.} We initialize the quantization parameters through grid search on 8 Pile~\citep{pile} samples with a 1024 sequence length. We minimize the layer outputs for fine-grained quantization (per-channel/per-head) and block outputs for per-tensor quantization.

\textbf{Fine-Tuning Setting.} During fine-tuning, we optimize block output mean square error following existing works~\citep{omniquant,efficientqat}. The dataset for fine-tuning consists of 512 samples from Pile with a 1024 context length. The learning rates for quantization parameters (step sizes) and full-precision weights are set to 5e-5 and 5e-6, respectively, and to 2e-5 and 2e-6 for Llama-3-70B(-Instruct) models. The fine-tuning batch size is set to 4, and the number of epochs is set to 10 for W4A8KV4 and 20 for W4A4KV4.

\subsection{Comparison Results}

\textbf{Results on W4A4KV4.} Table~\ref{tab:w4a4} presents the comparison results for W4A4KV4. \methodshort consistently outperforms existing methods. For example, under the same dynamic quantization setting on Llama-3-8B, PrefixQuant-O1 achieves a $1.12$ WikiText perplexity improvement and $+4.18$ points accuracy over DuQuant. Additionally, the more efficient PrefixQuant-O2 for static quantization also surpasses DuQuant, with a $0.71$ perplexity improvement and $+3.95$ points accuracy.

\textbf{Results on W4A8KV8.} Table~\ref{tab:w4a8} shows the comparison results for W4A8KV8. PrefixQuant-O1 and PrefixQuant-O2 outperform both QoQ and QuaRot across most models. For instance, PrefixQuant-O1 surpasses QoQ~\citep{qserve} by $0.31$ perplexity and $+1.22$ points accuracy on Llama-3-8B. Similarly, PrefixQuant-O2 maintains performance benefits with a $0.28$ perplexity improvement and $+1.11$ points accuracy.

\textbf{Results on more models.} The results in Table~\ref{tab:other_models} demonstrate that \methodshort consistently achieves excellent performance on other models such as Mistral-7b-v0.3 and Qwen-2-7B, as well as instruction-tuned models like Llama-3-\{7B,70B\}-Instruct.

\textbf{Results on MMLU.} Table~\ref{tab:mmlu_results} presents the comparison results on zero-shot MMLU using Llama-3-8B. PrefixQuant-O1 and PrefixQuant-O2 outperform SpinQuant by $+2.24$ and $+0.95$ accuracy, respectively, in W4A8KV4 quantization. The performance advantage is even more pronounced in W4A4KV4 quantization, with improvements of $+4.07$ and $+2.72$ accuracy, respectively.


\subsection{Ablation Studies}
We analyze the effects of various quantization techniques implemented in \methodshort. These techniques are applied incrementally, and the WikiText2 perplexity results are presented in Table~\ref{tab:ablation_techniques}.
The analysis begins with round-to-nearest (RTN) quantization on Llama-3-8B, incorporating Hadamard rotation and grid search initialization. Using W4A4KV4 as an example, we observe that introducing prefixed outliers significantly improves performance. Specifically, perplexity decreases from $11.70$ to $7.53$ for PrefixQuant-O1 and from $141.02$ to $7.93$ for PrefixQuant-O2. These improvements result not only from mitigating information loss caused by outlier tokens but also from enabling more accurate quantization parameter selection during grid searches initialization by isolating extremely large outliers (e.g., values exceeding $1e3$) in activations. Additionally, block-wise fine-tuning further enhances performance, reducing perplexity by $0.30$ for PrefixQuant-O1 and by $0.52$ for PrefixQuant-O2 in W4A4KV4 quantization.
Additional ablation results, including analyses of the training dataset, training epochs, dynamic quantization, the number of prefixed tokens, and the content of prefixed tokens, are provided in Sec.~\ref{sec:more_ablation} of the Appendix.

\begin{figure}[!h]
    \centering
    \setlength\tabcolsep{3pt}
    \renewcommand{\arraystretch}{0.5}
    \caption{Inference speedup of W4A4 Llama-2-7B model over the FP16 model on RTX 3090 GPU. For prefilling, we report the latency to deal with 2048 input tokens. For decoding, we report the token generation speed of generate 256 new tokens with 2048 prefillinng length. }
    \begin{tabular}{lll}
    \toprule
    \multirow{2}{*}{Method} & prefilling & decoding \\
    & (ms) & (token/s) \\
    \midrule
    FP16 & 489 & 43 \\
    \cdashline{1-3}
    PrefixQuant-O1 (W4A4)& 183 (\bf{2.67x}) & 91 (\textbf{2.11x}) \\
    PrefixQuant-O2 (W4A4) & 178 (\textbf{2.74x}) & 93 (\textbf{2.16x}) \\
    \bottomrule
    \end{tabular}
    \label{tab:e2e_prefilling}
\end{figure}

\subsection{Inference Speed}
In this section, we evaluate the end-to-end inference speed of PrefixQuant under the W4A4 quantization scenario. KV quantization is not considered because it reduces memory usage at the cost of increased computation overhead and only provides speedup with large batch sizes~\citep{kivi}. As shown in Table~\ref{tab:e2e_prefilling}, PrefixQuant achieves an approximate $2.7\times$ speedup in prefilling and a $2.1\times$ speedup in decoding compared to the FP16 model.
\section{Conclusion}
In this paper, we propose PrefixQuant, which provides a comprehensive exploration of outlier tokens and introduces an efficient and effective method to handle them by prefixing these tokens in the KV cache. Additionally, we design new trainable parameters for activation quantization to minimize quantization error. The proposed PrefixQuant method achieves excellent performance across various models, quantization precisions, and granularities. The simplicity and broad applicability of \methodshort make it a promising direction for future research on LLM compression and optimization.
\section{Impact Statement}
This paper presents work whose goal is to advance the field of Machine Learning. There are many potential societal consequences of our work, none which we feel must be specifically highlighted here.


\bibliography{example_paper}
\bibliographystyle{icml2025}

\newpage
\appendix
\onecolumn


\section*{Overview of Appendix}
We detailed the content of Appendix here:
\begin{itemize}
    \item Sec~\ref{sec:setting_and_results_source} details results sources of comparison methods. 
    \item Sec.~\ref{sec:details_of_rotation} illustrates the detailed image of hadamaed rotation within a transformer block.
    \item  Sec.~\ref{sec:quantization_time} details the quantization time of \methodshort. 
    \item Sec.~\ref{sec:more_ablation} gives more ablation studies of \methodshort, including the fine-tuning dataset, training epoch, and number of prefixed tokens.
    \item Sec.~\ref{sec:long_context} offers the comparison results in long-context scnarios. 
    \item Sec.~\ref{sec:weight-only-quantization} demonstrates that proposed \methodshort can also play as a plug-in to enhance the performance of existing weight-only quantization methods.
    \item Sec.~\ref{sec:full_results} presents the detailed accuracy number of each zero-shot task, and provide more results of \methodshort on Mistral-v0.3-7B, Qwen-2-7B, and Llama-3-\{8B,70B\}-Instruct.
    \item Sec.~\ref{sec:more_visualization} illustrate more visualization of inputs of linear layer and $\mathbf{Q}$/$\mathbf{K}$/$\mathbf{V}$ on more models, including Llama-3-\{8B,70B\}, Mistral-7B-v0.3, Qwen-2-7B.
\end{itemize}

\section{Results Sources of Comparison Methods}~\label{sec:setting_and_results_source}

We compare our proposed \methodshort with several other methods: QuaRot~\citep{quarot}, Atom~\citep{atom}, QoQ~\citep{qserve}, SmoothQuant~\citep{smoothquant}, SpinQuant~\citep{spinquant}, and EfficientQAT~\citep{efficientqat}. The data for our comparisons either come directly from the official publications of these methods, from other papers, or from our own reproduction of the methods. The source of the data for each method is outlined as follows:
\begin{itemize}
    \item \textbf{QuaRot}: We present the performance of QuaRot using the Pile calibration dataset. The results for Llama-2 models with W4A4KV4 come from QoQ~\citep{qserve}, while the rest are reproduced using the official open-source code.
    \item \textbf{DuQuant}: We reproduce DuQuant with Pild calibration dataset through their official open-source code. Note that we change the evaluation toolbox to lm-eval v0.4.2 for more accurate evaluation.
    \item \textbf{Atom}: We present the performance of Atom using the Pile calibration dataset. The results are sourced from QoQ~\citep{qserve}.
    \item \textbf{QoQ}: We present the performance of QoQ using the Pile calibration dataset. The results for Llama-2 come from QoQ~\citep{qserve}, and the Llama-3 results are reproduced using the official open-source code.
    \item \textbf{SmoothQuant}: We present the performance of SmoothQuant using the Pile calibration dataset. All results are reproduced using the open-source code from QoQ~\citep{qserve}.
    \item \textbf{SpinQuant}: All results are reproduced using the official open-source code and the pre-trained rotation matrix. Note that SpinQuant directly trains on the WikiText2 dataset.
    \item \textbf{EfficientQAT}: All results are reproduced using the official open-source code and the pre-quantized models.
\end{itemize}

\section{Details of Rotation}\label{sec:details_of_rotation}
\begin{figure}[!h]
    \centering
    \includegraphics[width=0.9\linewidth]{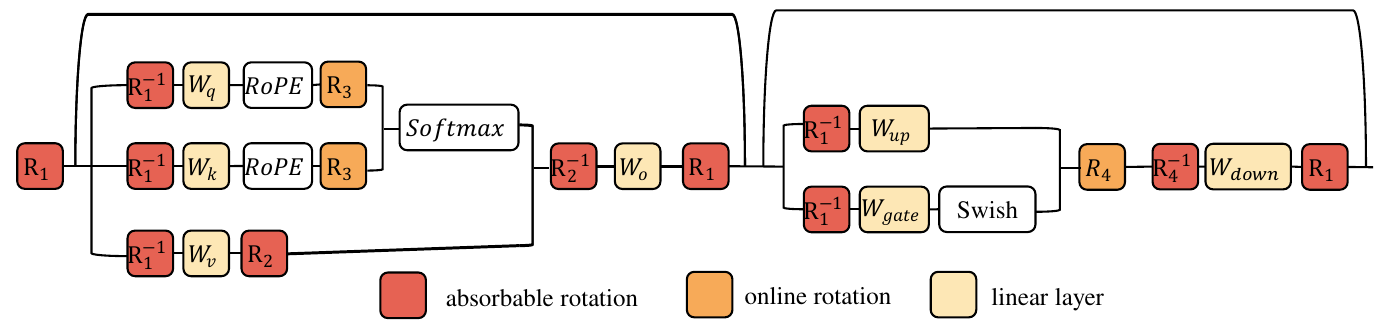}
    \caption{Illustrate of hadamard rotation within a transformer block of Llama~\citep{llama} model.}
    \label{fig:rotation}
\end{figure}

Hadamard rotation~\citep{quarot,spinquant} redistributes outlier channels across all channels, achieving uniform distribution within each token. The Hadamard matrix $\mathbf{H}$ is an orthogonal matrix with $\mathbf{H}\mathbf{H}^{T} = \mathbf{I}$, and its entries are $\{ +1, -1\}$ at the same scale.
Hadamard rotation can be applied to all activations and use inverse rotation on corresponding weights to maintain computational invariance~\citep{slicegpt}. Specifically, the rotation includes absorbable and online rotations. As shown in Figure~\ref{fig:rotation}, we follow SpinQuant~\citep{spinquant} to set $R1$, $R2$, $R3$ and $R4$ rotations, details as follows.

\textbf{Absorbable Rotation.} Hadamard rotation of activation can be absorbed into the previous linear layer if there is no intervening non-linear operation. Thus, the rotation of input activations for q/k/v/gate/up\_proj ($R_1$) and head-wise rotation for o\_proj input activations ($R_2$) can be fully absorbed without adding computation during inference.

\textbf{Online Rotation.} Some rotations must be executed online, including output activations of q\_proj and k\_proj after RoPE~\citep{rope} ($R_3$), and the input activation of down\_proj ($R_4$). These online rotations are efficiently implemented using the Walsh-Hadamard transform without significant overhead.

If not specifically mentioned, we activate all rotation ($R_1$, $R_2$, $R_3$ and $R_4$) in weight-activation quantization scenes, and only activate absorbable rotation ($R_1$ and $R_2$) in weight-only quantization.

\section{Quantization Time}~\label{sec:quantization_time}
Table~\ref{tab:quantization_time} shows the quantization time for \methodshort. \methodshort identifies prefixed tokens quickly, taking only 0.2 minutes for Llama-3-8B and 1 minute for Llama-3-70B. In contrast, the recent CushionCache~\citep{cushioncache} requires 12 hours for the same task on Llama-3-8B. Additionally, the grid-search initialization is efficient, taking 0.7 minutes for Llama-3-8B and 12 minutes for Llama-3-70B. 
Experiments in Tables~\ref{tab:w4a4} and \ref{tab:w4a8} demonstrate that \methodshort, even without fine-tuning, outperforms previous methods~\citep{qserve,quarot}. Fine-tuning requires more time, taking 2.2 hours for Llama-3-8B and 17 hours for Llama-3-70B, but it can successfully enhances the potential of low-bit quantization.
\begin{table}[h]
    \centering
    \caption{The quantization time of PrefixQuant on single NVIDIA-A100-80GB GPU. Fine-tuning indicates the time of 20 fine-tuning epochs of W4A4KV4.}
    \label{tab:quantization_time}
    \begin{tabular}{cccc}
    \toprule
    \bf{Model} & \bf{Find Prefixed Outliers} & \bf{Grid-search initialization} & \bf{Fine-tuning} \\ 
    \midrule
    Llama-3-8B & 0.2 m & 0.7 m & 2.2 h \\
    Llama-3-70B & 1 m & 12 m & 17 h \\
    \bottomrule
    \end{tabular}
\end{table}

\section{More Ablation Results}\label{sec:more_ablation}

\begin{table}[!h]
\renewcommand{\tabcolsep}{1pt}
\centering
    \caption{Ablation studies on calibration dataset, including (a) Dataset type, (b) Training sequence length and (c) Total training tokens. ``N'' indicates number of training samples, and ``S'' is the length of each samples. The model used here is Llama-3-8B with W4A4KV4 (PrefixQuant-O2) quantization. Our default settings are marked in {\setlength{\fboxsep}{2pt}\colorbox{mygray}{{gray}}}.}
\begin{subtable}[b]{0.30\linewidth}
    \centering
    \subcaption{Dataset}
    \label{tab:ablation_dataset}
    \begin{tabular}{cc}
    \toprule
    \bf{Dataset} & \bf{Wiki PPL} \\
    \midrule
    C4 & 7.60 \\
    RedPajama & 7.49  \\
    \cellcolor{mygray}Pile & \textbf{7.42}\\
    \bottomrule
    \end{tabular}
\end{subtable}
\begin{subtable}[b]{0.30\linewidth}
    \centering
    \subcaption{Sequence length}
    \label{tab:ablation_sequence_length}
    \begin{tabular}{cc}
    \toprule
    \bf{N $\times$ S} & \bf{Wiki PPL} \\
    \midrule
    256 $\times$2048 & 7.65  \\
    \cellcolor{mygray}512$\times$1024 & \textbf{7.42} \\
    1024$\times$512 & 7.65 \\
    \bottomrule
    \end{tabular}
\end{subtable}
\begin{subtable}[b]{0.30\linewidth}
    \centering
    \subcaption{Total token number}
    \label{tab:ablation_token_number}
    \begin{tabular}{cc}
    \toprule
    \bf{N $\times$ S} & \bf{Wiki PPL}  \\
    \midrule
    256 $\times$1024 & 7.46 \\
    \cellcolor{mygray}512 $\times$1024 & 7.42  \\
    1024$\times$1024 & \textbf{7.41} \\
    \bottomrule
    \end{tabular}
\end{subtable}

\end{table}
\begin{table}[!ht]
    \centering
    \caption{Ablation study about training epochs. The model used here is Llama-3-8B with PrefixQuant-O2, and WikiText2 perplexity with 2048 context length is reported. Our default settings are marked in {\setlength{\fboxsep}{2pt}\colorbox{mygray}{{gray}}}.}
    \label{tab:ablation_epoch}
    \begin{tabular}{ccc}
    \toprule
    \bf{Epochs} & \bf{W4A8KV4} & \bf{W4A4KV4} \\
    \midrule
    0 (w/o FT) & 6.90 & 7.93 \\
    5 & 6.66 & 7.53\\
    10 & \cellcolor{mygray}\textbf{6.63} & 7.47 \\
    20 & 6.63 & \cellcolor{mygray}7.42 \\
    30 & 6.63 & \textbf{7.41} \\
    \bottomrule
    \end{tabular}
    \label{tab:my_label}
\end{table}
\textbf{Fine-tuning Datasets.}
Table~\ref{tab:ablation_dataset} shows results with different fine-tuning datasets, including C4~\citep{c4}, RedPajama~\citep{redpajama}, and Pile~\citep{pile}. We find that Pile achieves the best performance. Additionally, we ablate the sequence length of each training sample and the total training tokens. Table~\ref{tab:ablation_sequence_length} shows that a sequence length of 1024 achieves the best performance. Table~\ref{tab:ablation_token_number} demonstrates that fine-tuning on $512 \times 1024$ tokens achieves satisfactory performance, with further increases in training samples only marginally improving performance.
Note that the optimal token number for fine-tuning datasets may change with quantization precision. Generally, lower precision requires more training data. For example, EfficientQAT shows that $4096 \times 2048$ tokens are needed for W2A16 quantization, while our paper shows that only $512 \times 1024$ tokens are needed for W4A4 quantization.

\textbf{Training Epochs.}
Table~\ref{tab:ablation_epoch} demonstrates that 10 and 20 epochs are sufficient for the convergence of fine-tuning on W4A8KV4 and W4A4KV4.


\begin{table}[!t]
    \centering
    \caption{Ablation study about the number of prefixed tokens. WikiText2 perplexity with 2048 context length and W4A4KV4 (PrefixQuant-O2) quantization is reported. Number $n$ indicates the first $n$ tokens in Table~\ref{tab:prefixed_token} are set as the prefixed tokens.}
    \begin{tabular}{ccccccc}
    \toprule
    \bf{Model} & \bf{Method} & \bf{0} & \bf{1} & \bf{2} & \bf{3} & \bf{4} \\
    \midrule
    Llama-2-7B & PrefixQuant w/o FT &  333.52 &  74.37 & 6.21 & 6.22 & -\\
    Llama-2-7B & PrefixQuant & 17.63 & 10.71 & \textbf{6.01} & \textbf{6.01} \\
    \cdashline{1-7}
    Mistral-7B-v0.3 & PrefixQuant w/o FT & 90.02 & 6.12 & 5.84 & 6.43 & 5.89 \\
    Mistral-7B-v0.3 & PrefixQuant & 15.97 & 7.08 & 5.83 & 5.95 & \textbf{5.79} \\
    \bottomrule
    \end{tabular}
    \label{tab:ablation_prefixed_number}
\end{table}
\begin{table}[!ht]
    \centering
    \caption{Ablation study about the content of prefixed tokens. WikiText2 perplexity with 2048 context length and W4A4KV4 (PrefixQuant-O2) quantization is reported. ``default'' refers to the prefixed tokens obtained through the proposed method. ``random'' represents the average performance of 10 times with randomly selected prefixed tokens.}
    \begin{tabular}{cclc}
    \toprule
    Model & Type & Prefixed & Wiki PPL (PrefixQuant w/o FT) \\
    \midrule
    Llama-2-7B & default &.\textbackslash n[BOS] & 6.22 \\
    Llama-2-7B & only highest frequency & ... & 12.07 \\
    Llama-2-7B & random & - & 66.51 \\
    \cdashline{1-4}
    Mistral-7B-v0.3 & default & \textbackslash n.to[BOS] & 5.89\\
    Mistral-7B-v0.3 & only highest frequency & \textbackslash n\textbackslash n\textbackslash n\textbackslash n & 6.23 \\
    Mistral-7B-v0.3 & random & - & 80.05 \\
    \bottomrule
    \end{tabular}
    \label{tab:ablation_content}
\end{table}

\textbf{Number of Prefixed Tokens.} In Sec.~\ref{sec:prefixed_outliers}, we determine the number of prefixed tokens by calculating the average number of outlier tokens and adding an additional [BOS] token. Table~\ref{tab:prefixed_token} illustrates the specific number and content of these tokens. We use Llama-2-7B (3 outlier tokens) and Mistral-7B-v0.3 (4 outlier tokens) to study the impact of the number of prefixed tokens. Table~\ref{tab:ablation_prefixed_number} shows that the adaptively calculated number of prefixed tokens achieves the best performance. Notably, for models like Llama-2-7B, using 2 prefixed tokens without the additional [BOS] token also yields excellent performance. For consistency and simplicity, we include the [BOS] token in the prefixed tokens in our experiments.

\textbf{Content of Prefixed Tokens.} \methodshort determines the number of outlier tokens $N$ and designates the top-$N$ high-frequency outlier tokens as prefixes in the KV cache. Table~\ref{tab:ablation_content} examines various prefixed tokens with the same token count. The results show that using the top-$N$ high-frequency tokens as prefixed tokens significantly outperforms using only the highest-frequency or randomly selected tokens.

\begin{table}[!t]
\renewcommand{\arraystretch}{0.4}
    \centering
    \caption{Comparisons in long-context scenario of Llama-3-8B. We report the WikiText2 perplexity with context length 8192. We do not report SpinQuant results because it overfits to WikiText2 datasets. }
    \begin{tabular}{ccc}
    \toprule
    Method & Precision & PPL. \\
    \midrule
    - & FP16 & 5.54 \\
    \midrule
    QuaRot & W4A8KV4 & 6.79 \\
    DuQuant & W4A8KV4 & 6.19 \\
     \cellcolor{mylightblue}PrefixQuant-O1 &  \cellcolor{mylightblue}W4A8KV4 &  \cellcolor{mylightblue}\textbf{5.94} \\
     \cellcolor{mylightblue}PrefixQuant-O2 &  \cellcolor{mylightblue}W4A8KV4 &   \cellcolor{mylightblue} 6.04 \\
    \hline
    QuaRot & W4A4KV4 & 8.41 \\
    DuQuant & W4A4KV4 & 7.27 \\
     \cellcolor{mylightblue}PrefixQuant-O1 &  \cellcolor{mylightblue}W4A4KV4 &  \cellcolor{mylightblue}\textbf{6.58} \\
     \cellcolor{mylightblue}PrefixQuant-O2 &  \cellcolor{mylightblue}W4A4KV4 & \cellcolor{mylightblue}6.82 \\
    \bottomrule
    \end{tabular}
    \label{tab:long_context_comparisons}
\end{table}
\section{Comparisons in Long-Context Scenarios}\label{sec:long_context}
PrefixQuant-O1 uses a shared clipping factor for each layer, while PrefixQuant-O2 further shares a scaling factor across the entire tensor. A longer input context implies that more activations share the same clipping or scaling factor. This raises a concern about whether PrefixQuant remains effective in long-context scenarios. Table~\ref{tab:long_context_comparisons} shows that PrefixQuant consistently outperforms existing methods at a context length of 8192, demonstrating the strong generalization ability of the proposed method.

\begin{table*}[!ht]
\vspace{-0.5em}
\footnotesize
\centering
\setlength\tabcolsep{2.0pt}
  \caption{Weight-only quantization results. ``g'' indicates group size for weight quantization. EfficientQAT only execute Block-AP and without E2E-QP for the fair comparisons in block-wise reconstruction scenario. We providing WikiText2 perplexity with 2048 context length and detailed zero-shot accuracy of weight-only quantization by \texttt{lm\_eval v0.4.2}. We report \texttt{acc} for WinoGrande and \texttt{acc\_norm} for HellaSwag, ArcC, ArcE, and PIQA.}\label{tab:full_acc_results_weight_only}
\begin{tabular}{lcc|ccccccc}
 \toprule
  \bf{Model} & \bf{Method} & \bf{Precision} & \bf{Wiki PPL} & \bf{WinoGrande} & \bf{HellaSwag}  & \bf{ArcC} & \bf{ArcE}  & \bf{PiQA} & \bf{Avg. Acc.}\\
  \midrule
  \multirow{5}{*}{3-8B} & Baseline  & FP16  & 6.14 & 72.61 & 79.17 & 53.41 & 77.69 & 80.69 & 72.71  \\
  \cdashline{2-10}
  & EfficientQAT & W3A16g128  & 7.34 & 70.48 & 75.09 & 51.37 & 77.9 & 79.16 & 70.80 \\
  & \cellcolor{mylightblue}PrefixQuant & \cellcolor{mylightblue}W3A16g128  & \cellcolor{mylightblue}7.17 & \cellcolor{mylightblue}72.38 & \cellcolor{mylightblue}76.54 & \cellcolor{mylightblue}52.65 & \cellcolor{mylightblue}78.37 & \cellcolor{mylightblue}80.58 & \cellcolor{mylightblue}72.10 \\
  \cdashline{2-10}
  & EfficientQAT & W2A16g128  & 13.55 & 62.04 & 62.49 & 36.6 & 60.44 & 73.18 & 58.95 \\
  & \cellcolor{mylightblue}PrefixQuant & \cellcolor{mylightblue}W2A16g128  & \cellcolor{mylightblue}11.97 & \cellcolor{mylightblue}66.22 & \cellcolor{mylightblue}66.54 & \cellcolor{mylightblue}41.81 & \cellcolor{mylightblue}69.61 & \cellcolor{mylightblue}75.84 & \cellcolor{mylightblue}64.00 \\
  \midrule
  \multirow{5}{*}{3-70B} & Baseline  & FP16 & 2.85 & 80.51 & 84.9 & 64.33 & 85.9 & 84.49 & 80.03\\
  \cdashline{2-10}
  & EfficientQAT & W3A16g128  & 4.89 & 78.77 & 83.74 & 55.03 & 78.66 & 82.05 & 75.65 \\
  & \cellcolor{mylightblue}PrefixQuant & \cellcolor{mylightblue}W3A16g128  & \cellcolor{mylightblue}4.79 & \cellcolor{mylightblue}78.22 & \cellcolor{mylightblue}84.03 & \cellcolor{mylightblue}60.15 & \cellcolor{mylightblue}83.00 & \cellcolor{mylightblue}83.35 & \cellcolor{mylightblue}77.75  \\
  \cdashline{2-10}
  & EfficientQAT & W2A16g128  & 16.79 & 66.14 & 73.01 & 48.21 & 73.57 & 78.45 & 67.88 \\
  & \cellcolor{mylightblue}PrefixQuant & \cellcolor{mylightblue}W2A16g128  & \cellcolor{mylightblue}11.01 & \cellcolor{mylightblue}72.3 & \cellcolor{mylightblue}78.55 & \cellcolor{mylightblue}53.67 & \cellcolor{mylightblue}77.9 & \cellcolor{mylightblue}80.63 & \cellcolor{mylightblue}72.61 \\
 \bottomrule
\end{tabular}

\end{table*}

\section{Extend to Weight-only Quantization}\label{sec:weight-only-quantization}
In addition to static activation quantization, setting prefixed outliers in the KV-cache improves training stability~\citep{ternaryllm} and reduces information loss from outlier tokens, can also enhancing weight-only quantization performance. To verify this, we compare \methodshort with the recent state-of-the-art weight-only quantization method, EfficientQAT~\citep{efficientqat}, in a block-wise fine-tuning scenario. Following EfficientQAT, we use 4096 RedPajama~\citep{redpajama} with a 2048 context length to train for 2 epochs. The learning rates for quantization parameters and full-precision weights are set to 5e-5 and 5e-6, except for W2A16g128 Llama-3-8B, where they are 1e-4 and 2e-5, respectively. As shown in Table~\ref{tab:full_acc_results_weight_only}, \methodshort significantly surpasses EfficientQAT with $+5.05$ and $+4.73$ points in average accuracy on W2A16g128 Llama-3-8B and Llama-3-70B, respectively.

\section{Full Results of Weight-Activation quantization}~\label{sec:full_results}

\begin{table}[h]
    \centering
    \caption{W8A8 performance comparisons with other methods that also set prefixed tokens in KV cache.}
    \begin{tabular}{cccc}
    \toprule
    Model & Method & Activation Quant & Wiki PPL \\
    \midrule
    \multirow{3}{*}{2-7B} & QFeP & per-tensor dynamic & 5.75  \\
    & CushionCache & per-tensor static & 5.87 \\
    & \cellcolor{mylightblue}PrefixQuant-O2 & \cellcolor{mylightblue}per-tensor static & \cellcolor{mylightblue}\cellcolor{mylightblue}\bf{5.48} \\
    \cdashline{1-4}
    \multirow{2}{*}{2-13B} & QFeP & per-tensor dynamic & 6.00 \\
    & \cellcolor{mylightblue}PrefixQuant-O2 & \cellcolor{mylightblue}per-tensor static & \cellcolor{mylightblue}\bf{4.89} \\
    \cdashline{1-4}
    \multirow{2}{*}{2-70B}& QFeP & per-tensor dynamic & 6.01 \\
    & \cellcolor{mylightblue}PrefixQuant-O2 & \cellcolor{mylightblue}per-tensor static & \cellcolor{mylightblue}\bf{3.39} \\
    \cdashline{1-4}
    \multirow{2}{*}{3-8B}& CushionCache & per-tensor static & 7.37 \\
    & \cellcolor{mylightblue}PrefixQuant-O2 & \cellcolor{mylightblue}per-tensor static &  \cellcolor{mylightblue}\bf{6.17} \\
    \bottomrule
    \end{tabular}
    \label{tab:prefix_comparisons}
\end{table}
\subsection{Comparisons with Related Works}
CushionCache~\citep{cushioncache} and QFeP~\citep{qfep} also set prefixed tokens in the KV cache to reduce outliers. However, they experience significant performance degradation even with W8A8 quantization. Table~\ref{tab:prefix_comparisons} shows that \methodshort outperforms QFeP by 2.62 perplexity on Llama-2-70B and surpasses CushionCache by 1.20 perplexity on Llama-3-8B.

\subsection{Detailed Accuracy Results}
In the main paper, we present the average accuracy of five common reasoning tasks for brevity. Here, we provide detailed results for each task in Table~\ref{tab:full_acc_results}.

\subsection{Results on More Models}\label{sec:results_on_more_models}
Table~\ref{tab:other_models} shows the effectiveness of the proposed \methodshort in other models, including Mistral-v0.3-7B and Qwen-2-7B. It also includes instruction-tuned models such as Llama-3-\{8B,70B\}-Instruct.

\begin{table*}[!ht]
\vspace{-0.5em}
\footnotesize
\centering
\setlength\tabcolsep{2.37pt}
\renewcommand{\arraystretch}{0.4}
  \caption{Continuation of Table~\ref{tab:w4a4} and Table~\ref{tab:w4a8}, providing detailed zero-shot accuracy of weight-activation quantization of Llama models by \texttt{lm\_eval v0.4.2}. We report \texttt{acc} for WinoGrande and \texttt{acc\_norm} for HellaSwag, ArcC, ArcE, and PIQA.).}\label{tab:full_acc_results}
\begin{tabular}{lcc|cccccc}
 \toprule
  \bf{Model} & \bf{Method} & \bf{Precision} & \bf{WinoGrande} & \bf{HellaSwag}  & \bf{ArcC} & \bf{ArcE}  & \bf{PiQA} & \bf{Avg. Acc.}\\
  \midrule
  \multirow{14}{*}{2-7B} & Baseline  & FP16   & 69.22 & 76.00 & 46.25 & 74.62 & 79.11 & 69.04   \\
  \cdashline{2-9}
  & Atom & W4A4KV4  & 62.75 & 69.37 & 38.40 & 52.99 & 75.14 & 59.73\\
  & QuaRot & W4A4KV4  & 64.40 & 72.3 & 41.47 & 68.06 & 76.17 & 64.48 \\
  & DuQuant & W4A4KV4  & 67.09 & 72.53 & 43.26 & 71.38 & 76.99 & 66.25  \\
  & SpinQuant & W4A4KV4  & 66.54 & 73.15 & 41.64 & 69.32 & 76.12 & 65.35  \\
  & \cellcolor{mylightblue}PrefixQuant-O1 w/o FT & \cellcolor{mylightblue}W4A4KV4  & \cellcolor{mylightblue} 66.85 & \cellcolor{mylightblue} 74.27 & \cellcolor{mylightblue} 43.86  & \cellcolor{mylightblue} 72.35  & \cellcolor{mylightblue} 77.97 & \cellcolor{mylightblue} \textbf{67.06}  \\
  & \cellcolor{mylightblue}PrefixQuant-O1 & \cellcolor{mylightblue}W4A4KV4  & \cellcolor{mylightblue} 67.48 & \cellcolor{mylightblue} 73.77 & \cellcolor{mylightblue} 43.17  & \cellcolor{mylightblue} 71.3  & \cellcolor{mylightblue} 77.97 & \cellcolor{mylightblue} 66.74  \\
  & \cellcolor{mylightblue}PrefixQuant-O2 w/o FT & \cellcolor{mylightblue}W4A4KV4  & \cellcolor{mylightblue}67.80 & \cellcolor{mylightblue}73.75 & \cellcolor{mylightblue}43.94 & \cellcolor{mylightblue}71.51 & \cellcolor{mylightblue}77.2 & \cellcolor{mylightblue}66.84 \\
  & \cellcolor{mylightblue}PrefixQuant-O2 & \cellcolor{mylightblue}W4A4KV4  & \cellcolor{mylightblue}66.54 & \cellcolor{mylightblue}73.42 & \cellcolor{mylightblue}43.09 & \cellcolor{mylightblue}71.17 & \cellcolor{mylightblue}77.64 & \cellcolor{mylightblue}66.37 \\
  \cdashline{2-9}
  & QoQ & W4A8KV4 & 68.03 & 74.00 & 43.60 & 72.81 & 77.64 & 67.22 \\
  & QuaRot & W4A8KV4  & 66.77 & 74.56 & 43.86 & 72.39 & 77.97 & 67.11 \\
  & \cellcolor{mylightblue}PrefixQuant-O1 w/o FT & \cellcolor{mylightblue}W4A8KV8  & \cellcolor{mylightblue} 69.53 & \cellcolor{mylightblue} 75.49 & \cellcolor{mylightblue} 44.8  & \cellcolor{mylightblue} 73.32  & \cellcolor{mylightblue} 77.64 & \cellcolor{mylightblue} \textbf{68.16}  \\
  & \cellcolor{mylightblue}PrefixQuant-O1 & \cellcolor{mylightblue}W4A8KV8  & \cellcolor{mylightblue} 69.69 & \cellcolor{mylightblue} 75.3 & \cellcolor{mylightblue} 44.28  & \cellcolor{mylightblue} 73.19  & \cellcolor{mylightblue} 77.75 & \cellcolor{mylightblue} 68.04  \\
  & \cellcolor{mylightblue}PrefixQuant-O2 w/o FT & \cellcolor{mylightblue}W4A8KV4  & \cellcolor{mylightblue}69.14 & \cellcolor{mylightblue}75.12 & \cellcolor{mylightblue}44.45 & \cellcolor{mylightblue}73.06 & \cellcolor{mylightblue}77.53 & \cellcolor{mylightblue}67.86 \\
  & \cellcolor{mylightblue}PrefixQuant-O2 & \cellcolor{mylightblue}W4A8KV4  & \cellcolor{mylightblue}69.06 & \cellcolor{mylightblue}75.25 & \cellcolor{mylightblue}44.8 & \cellcolor{mylightblue}73.19 & \cellcolor{mylightblue}78.13 & \cellcolor{mylightblue}68.09 \\
 \midrule
  \multirow{14}{*}{2-13B} & Baseline  & FP16  & 72.22 & 79.37 & 49.06 & 77.48 & 80.52 & 71.73  \\
  \cdashline{2-9}
  & Atom & W4A4KV4 & 67.40 & 73.84 & 42.32 & 57.49 & 76.50 & 63.51 \\
  & QuaRot & W4A4KV4  & 67.88 & 75.28 & 45.65 & 72.35 & 77.48 & 67.73 \\
  & DuQuant & W4A4KV4  & 68.9 & 76.65 & 47.7 & 74.24 & 78.18 & 69.13  \\
  & SpinQuant & W4A4KV4  & 67.88 & 77.01 & 46.76 & 75.97 & 78.56 & 69.24 \\
    & \cellcolor{mylightblue}PrefixQuant-O1 w/o FT & \cellcolor{mylightblue}W4A4KV4  & \cellcolor{mylightblue} 72.38 & \cellcolor{mylightblue} 76.92 & \cellcolor{mylightblue} 47.7  & \cellcolor{mylightblue} 75.8  & \cellcolor{mylightblue} 79.43 & \cellcolor{mylightblue} \textbf{70.45}  \\
  & \cellcolor{mylightblue}PrefixQuant-O1 & \cellcolor{mylightblue}W4A4KV4  & \cellcolor{mylightblue} 71.98 & \cellcolor{mylightblue} 76.3 & \cellcolor{mylightblue} 46.59  & \cellcolor{mylightblue} 76.43  & \cellcolor{mylightblue} 78.94 & \cellcolor{mylightblue} 70.05  \\
  & \cellcolor{mylightblue}PrefixQuant-O2 w/o FT & \cellcolor{mylightblue}W4A4KV4  & \cellcolor{mylightblue}72.06 & \cellcolor{mylightblue}76.54 & \cellcolor{mylightblue}46.67 & \cellcolor{mylightblue}75.8 & \cellcolor{mylightblue}78.51 & \cellcolor{mylightblue}69.92 \\
  & \cellcolor{mylightblue}PrefixQuant-O2 & \cellcolor{mylightblue}W4A4KV4  & \cellcolor{mylightblue}72.53 & \cellcolor{mylightblue}76.12 & \cellcolor{mylightblue}47.70 & \cellcolor{mylightblue}76.09 & \cellcolor{mylightblue}79.38 & \cellcolor{mylightblue}70.36 \\
  \cdashline{2-9}
  & QoQ & W4A8KV4 & 70.96 & 77.80 & 48.38 & 75.97 & 79.71 & 70.56\\
  & QuaRot & W4A8KV4 & 70.24 & 78.21 & 47.01 & 74.49 & 79.87 & 69.96 \\
  & \cellcolor{mylightblue}PrefixQuant-O1 w/o FT & \cellcolor{mylightblue}W4A8KV8  & \cellcolor{mylightblue} 72.77 & \cellcolor{mylightblue} 77.6 & \cellcolor{mylightblue} 48.46  & \cellcolor{mylightblue} 77.36  & \cellcolor{mylightblue} 80.36 & \cellcolor{mylightblue} \textbf{71.31}  \\
  & \cellcolor{mylightblue}PrefixQuant-O1 & \cellcolor{mylightblue}W4A8KV8  & \cellcolor{mylightblue} 72.14 & \cellcolor{mylightblue} 77.71 & \cellcolor{mylightblue} 48.98  & \cellcolor{mylightblue} 76.89  & \cellcolor{mylightblue} 80.52 & \cellcolor{mylightblue} 71.25  \\
  & 
  \cellcolor{mylightblue}PrefixQuant-O2 w/o FT & \cellcolor{mylightblue}W4A8KV4  & \cellcolor{mylightblue}72.77 & \cellcolor{mylightblue}77.49 & \cellcolor{mylightblue}48.12 & \cellcolor{mylightblue}77.06 & \cellcolor{mylightblue}79.92 & \cellcolor{mylightblue}71.07 \\
  & \cellcolor{mylightblue}PrefixQuant-O2 & \cellcolor{mylightblue}W4A8KV4  & \cellcolor{mylightblue}72.77 & \cellcolor{mylightblue}77.54 & \cellcolor{mylightblue}48.72 & \cellcolor{mylightblue}76.81 & \cellcolor{mylightblue}80.41 & \cellcolor{mylightblue}71.25 \\
 \midrule
  \multirow{14}{*}{2-70B} & Baseline  & FP16 & 79.48 & 84.31 & 56.91 & 80.30 & 82.54 & 76.71 \\
  \cdashline{2-9}
  & Atom & W4A4KV4 & 74.27 & 79.06 & 46.08 & 58.25 & 79.92 & 67.52 \\
  & QuaRot & W4A4KV4 & 76.24 & 81.82 & 56.23 & 80.43 & 82.43 & 75.43 \\
  & DuQuant & W4A4KV4  & 75.45 & 81.95 & 55.03 & 79 & 82.32 & 74.75  \\
  & SpinQuant & W4A4KV4 & 75.85 & 82.36 & 56.31 & 79.17 & 81.61 & 75.19 \\
    & \cellcolor{mylightblue}PrefixQuant-O1 w/o FT & \cellcolor{mylightblue}W4A4KV4  & \cellcolor{mylightblue} 77.98 & \cellcolor{mylightblue} 81.38 & \cellcolor{mylightblue} 55.55  & \cellcolor{mylightblue} 78.7  & \cellcolor{mylightblue} 81.12 & \cellcolor{mylightblue} 74.95  \\
  & \cellcolor{mylightblue}PrefixQuant-O1 & \cellcolor{mylightblue}W4A4KV4  & \cellcolor{mylightblue} 78.77 & \cellcolor{mylightblue} 83.23 & \cellcolor{mylightblue} 56.48  & \cellcolor{mylightblue} 79.92  & \cellcolor{mylightblue} 82.75 & \cellcolor{mylightblue} \textbf{76.23}  \\
  & \cellcolor{mylightblue}PrefixQuant-O2 w/o FT & \cellcolor{mylightblue}W4A4KV4  & \cellcolor{mylightblue}75.45 & \cellcolor{mylightblue}80.51 & \cellcolor{mylightblue}52.3 & \cellcolor{mylightblue}77.06 & \cellcolor{mylightblue}81.12 & \cellcolor{mylightblue}73.29 \\
  & \cellcolor{mylightblue}PrefixQuant-O2 & \cellcolor{mylightblue}W4A4KV4  & \cellcolor{mylightblue}77.35 & \cellcolor{mylightblue}82.3 & \cellcolor{mylightblue}56.4 & \cellcolor{mylightblue}79.29 & \cellcolor{mylightblue}82.05 & \cellcolor{mylightblue}75.48 \\
  \cdashline{2-9}
  & QoQ & W4A8KV4 & 77.51 & 82.78 & 56.83 & 79.80 & 82.64 & 75.91 \\
  & QuaRot & W4A8KV4 & 77.03	& 83.30	& 57.08 & 81.27 & 	82.86	& 76.31 \\
  & \cellcolor{mylightblue}PrefixQuant-O1 w/o FT & \cellcolor{mylightblue}W4A8KV8  & \cellcolor{mylightblue} 78.06 & \cellcolor{mylightblue} 83.64 & \cellcolor{mylightblue} 55.12  & \cellcolor{mylightblue} 79.71  & \cellcolor{mylightblue} 82.15 & \cellcolor{mylightblue} 75.74  \\
  & \cellcolor{mylightblue}PrefixQuant-O1 & \cellcolor{mylightblue}W4A8KV8  & \cellcolor{mylightblue} 79.64 & \cellcolor{mylightblue} 83.97 & \cellcolor{mylightblue} 57.68  & \cellcolor{mylightblue} 80.18  & \cellcolor{mylightblue} 82.64 & \cellcolor{mylightblue} \textbf{76.82}  \\
  & \cellcolor{mylightblue}PrefixQuant-O2 w/o FT & \cellcolor{mylightblue}W4A8KV4  & \cellcolor{mylightblue}77.35 & \cellcolor{mylightblue}82.79 & \cellcolor{mylightblue}54.35 & \cellcolor{mylightblue}78.28 & \cellcolor{mylightblue}82.21 & \cellcolor{mylightblue}75.00 \\
  & \cellcolor{mylightblue}PrefixQuant-O2 & \cellcolor{mylightblue}W4A8KV4  & \cellcolor{mylightblue}79.08 & \cellcolor{mylightblue}83.56 & \cellcolor{mylightblue}57.42 & \cellcolor{mylightblue}80.39 & \cellcolor{mylightblue}82.05 & \cellcolor{mylightblue}76.50 \\
 \midrule
  \multirow{14}{*}{3-8B} & Baseline  & FP16  & 72.61 & 79.17 & 53.41 & 77.69 & 80.69 & 72.71 \\
  \cdashline{2-9}
  & QuaRot & W4A4KV4  & 65.98 & 72.38 & 44.45 & 67.3 & 75.63 & 65.15  \\
  & DuQuant & W4A4KV4  & 68.59 & 74.27 & 46.5 & 70.41 & 75.9 & 67.13  \\
  & SpinQuant & W4A4KV4  & 69.22 & 74.83 & 45.99 & 74.07 & 77.04 & 68.23 \\
    & \cellcolor{mylightblue}PrefixQuant-O1 w/o FT & \cellcolor{mylightblue}W4A4KV4  & \cellcolor{mylightblue} 70.32 & \cellcolor{mylightblue} 75.86 & \cellcolor{mylightblue} 48.38  & \cellcolor{mylightblue} 71.46  & \cellcolor{mylightblue} 77.58 & \cellcolor{mylightblue} 68.72  \\
  & \cellcolor{mylightblue}PrefixQuant-O1 & \cellcolor{mylightblue}W4A4KV4  & \cellcolor{mylightblue} 70.88 & \cellcolor{mylightblue} 75.95 & \cellcolor{mylightblue} 52.47  & \cellcolor{mylightblue} 78.7  & \cellcolor{mylightblue} 79.6 & \cellcolor{mylightblue} \textbf{71.52}  \\
  & \cellcolor{mylightblue}PrefixQuant-O2 w/o FT & \cellcolor{mylightblue}W4A4KV4  & \cellcolor{mylightblue}69.14 & \cellcolor{mylightblue}75.46 & \cellcolor{mylightblue}47.1 & \cellcolor{mylightblue}72.94 & \cellcolor{mylightblue}77.2 & \cellcolor{mylightblue}68.37 \\
  & \cellcolor{mylightblue}PrefixQuant-O2 & \cellcolor{mylightblue}W4A4KV4  & \cellcolor{mylightblue}71.9 & \cellcolor{mylightblue}75.44 & \cellcolor{mylightblue}50.68 & \cellcolor{mylightblue}78.32 & \cellcolor{mylightblue}79.05 & \cellcolor{mylightblue}71.08 \\
  \cdashline{2-9}
  & QoQ & W4A8KV4  & 73.4 & 77.23 & 50.87 & 75.59 & 79.65 & 71.35 \\
  & QuaRot & W4A8KV4  & 72.74 & 77.35 & 51.62 & 77.48 & 79.22 & 71.68 \\
  & \cellcolor{mylightblue}PrefixQuant-O1 w/o FT & \cellcolor{mylightblue}W4A8KV8  & \cellcolor{mylightblue} 71.74 & \cellcolor{mylightblue} 77.99 & \cellcolor{mylightblue} 50.17  & \cellcolor{mylightblue} 74.07  & \cellcolor{mylightblue} 79.22 & \cellcolor{mylightblue} 70.64  \\
  & \cellcolor{mylightblue}PrefixQuant-O1 & \cellcolor{mylightblue}W4A8KV8  & \cellcolor{mylightblue} 73.16 & \cellcolor{mylightblue} 77.95 & \cellcolor{mylightblue} 52.56  & \cellcolor{mylightblue} 79.17  & \cellcolor{mylightblue} 80.03 & \cellcolor{mylightblue} \textbf{72.57}  \\
  & \cellcolor{mylightblue}PrefixQuant-O2 w/o FT & \cellcolor{mylightblue}W4A8KV4  & \cellcolor{mylightblue}71.19 & \cellcolor{mylightblue}77.65 & \cellcolor{mylightblue}48.98 & \cellcolor{mylightblue}73.99 & \cellcolor{mylightblue}79.65 & \cellcolor{mylightblue}70.29 \\
  & \cellcolor{mylightblue}PrefixQuant-O2 & \cellcolor{mylightblue}W4A8KV4  & \cellcolor{mylightblue}72.53 & \cellcolor{mylightblue}77.97 & \cellcolor{mylightblue}52.65 & \cellcolor{mylightblue}79.25 & \cellcolor{mylightblue}79.92 & \cellcolor{mylightblue}72.46 \\
 \midrule
  \multirow{14}{*}{3-70B} & Baseline  & FP16  & 80.51 & 84.9 & 64.33 & 85.9 & 84.49 & 80.03  \\
  \cdashline{2-9}
  & QuaRot & W4A4KV4  & 68.51 & 76.75 & 47.01 & 72.31 & 77.37 & 68.39 \\
  & DuQuant & W4A4KV4  & 70.8 & 79.89 & 59.04 & 82.91 & 81.83 & 74.89  \\
  & SpinQuant & W4A4KV4  & 76.4 & 80.9 & 56 & 77.3 & 80.8 & 74.28  \\
    & \cellcolor{mylightblue}PrefixQuant-O1 w/o FT & \cellcolor{mylightblue}W4A4KV4  & \cellcolor{mylightblue} 77.74 & \cellcolor{mylightblue} 83.61 & \cellcolor{mylightblue} 58.19  & \cellcolor{mylightblue} 80.3  & \cellcolor{mylightblue} 82.43 & \cellcolor{mylightblue} 76.45  \\
  & \cellcolor{mylightblue}PrefixQuant-O1 & \cellcolor{mylightblue}W4A4KV4  & \cellcolor{mylightblue} 77.74 & \cellcolor{mylightblue} 84.06 & \cellcolor{mylightblue} 58.96  & \cellcolor{mylightblue} 81.31  & \cellcolor{mylightblue} 83.35 & \cellcolor{mylightblue} 77.08  \\
  & \cellcolor{mylightblue}PrefixQuant-O2 w/o FT & \cellcolor{mylightblue}W4A4KV4  & \cellcolor{mylightblue}77.43 & \cellcolor{mylightblue}83.48 & \cellcolor{mylightblue}58.87 & \cellcolor{mylightblue}79.88 & \cellcolor{mylightblue}82.32 & \cellcolor{mylightblue}76.40 \\
  & \cellcolor{mylightblue}PrefixQuant-O2 & \cellcolor{mylightblue}W4A4KV4  & \cellcolor{mylightblue}77.35 & \cellcolor{mylightblue}83.79 & \cellcolor{mylightblue}60.15 & \cellcolor{mylightblue}81.31 & \cellcolor{mylightblue}83.3 & \cellcolor{mylightblue}\textbf{77.18} \\
  \cdashline{2-9}
  & QoQ & W4A8KV4  & 80.11 & 83.7 & 61.01 & 82.79 & 83 & 78.12 \\
  & QuaRot & W4A8KV4  & 80.35 & 84.03 & 62.12 & 84.64 & 83.46 & 78.92 \\
  & \cellcolor{mylightblue}PrefixQuant-O1 w/o FT & \cellcolor{mylightblue}W4A8KV8  & \cellcolor{mylightblue} 78.14 & \cellcolor{mylightblue} 84.92 & \cellcolor{mylightblue} 59.73  & \cellcolor{mylightblue} 81.06  & \cellcolor{mylightblue} 83.79 & \cellcolor{mylightblue} 77.53  \\
  & \cellcolor{mylightblue}PrefixQuant-O1 & \cellcolor{mylightblue}W4A8KV8  & \cellcolor{mylightblue} 79.4 & \cellcolor{mylightblue} 85.03 & \cellcolor{mylightblue} 61.69  & \cellcolor{mylightblue} 81.9  & \cellcolor{mylightblue} 84.49 & \cellcolor{mylightblue} 78.50  \\
  & \cellcolor{mylightblue}PrefixQuant-O2 w/o FT & \cellcolor{mylightblue}W4A8KV4  & \cellcolor{mylightblue}79.23 & \cellcolor{mylightblue}84.71 & \cellcolor{mylightblue}59.39 & \cellcolor{mylightblue}81.57 & \cellcolor{mylightblue}84.22 & \cellcolor{mylightblue}77.82 \\
  & \cellcolor{mylightblue}PrefixQuant-O2 & \cellcolor{mylightblue}W4A8KV4  & \cellcolor{mylightblue}79.48 & \cellcolor{mylightblue}84.86 & \cellcolor{mylightblue}62.29 & \cellcolor{mylightblue}82.53 & \cellcolor{mylightblue}84.33 & \cellcolor{mylightblue}\textbf{78.70} \\
 \bottomrule
\end{tabular}

\end{table*}
\begin{table}[!ht]
    \centering
    \setlength\tabcolsep{1pt}
    \caption{Results of proposed \methodshort-O1/O2 on other models.}
    \begin{tabular}{ccccccccc}
    \toprule
    \bf{Model} & \bf{Precision} & \bf{Wiki PPL} & \bf{WinoGrande} & \bf{HellaSwag}  & \bf{ArcC} & \bf{ArcE}  & \bf{PiQA} & \bf{Avg. Acc.} \\
    \midrule
    \multirow{4}{*}{Mistral-v0.3-7B} & FP16 & 5.32  & 73.88 & 80.43 & 52.3 & 78.28 & 82.26 & 73.43\\
    \cdashline{2-9} 
    & W4A8KV4-O1 & 5.49 & 72.93 & 80.14 & 52.13 & 78.79 & 81.34 & 73.07  \\
    & W4A8KV4-O2 & 5.51  & 73.88 & 79.8 & 52.05 & 79.42 & 80.79 & 73.19 \\
    & W4A4KV4-O1 & 5.76 & 70.88 & 78.39 & 50 & 77.74 & 80.41 & 71.48  \\
    & W4A4KV4-O2 & 5.79  & 71.51 & 78.12 & 49.66 & 78.03 & 79.92 & 71.45 \\
    \midrule
    \multirow{4}{*}{Qwen-2-7B} & FP16 & 7.14  & 72.3 & 78.96 & 52.65 & 78.75 & 80.96 & 72.72 \\
    \cdashline{2-9} 
    & W4A8KV4-O1 &  7.36 & 72.53 & 77.75 & 50.77 & 76.77 & 80.41 & 71.65\\
    & W4A8KV4-O2 & 8.04  & 71.43 & 76.77 & 53.67 & 77.95 & 78.45 & 71.65 \\
    & W4A4KV4-O1 & 7.76 & 69.93 & 75.56 & 51.02 & 76.81 & 78.56 & 70.38  \\
    & W4A4KV4-O2 & 8.37  & 68.75 & 74.92 & 48.21 & 74.75 & 79.49 & 69.22 \\
    \midrule
    \multirow{4}{*}{Llama-3-8B-Instruct} & FP16 & 8.29  & 71.82 & 75.81 & 56.83 & 79.76 & 78.51 & 72.55 \\
    \cdashline{2-9} 
    & W4A8KV4-O1 & 8.73 & 70.8 & 74.29 & 54.18 & 76.94 & 78.78 & 71.00  \\
    & W4A8KV4-O2 & 8.74  & 70.17 & 74.6 & 54.44 & 77.65 & 77.97 & 70.97 \\
    & W4A4KV4-O1 &  \\
    & W4A4KV4-O2 & 8.96  & 69.53 & 74.66 & 52.65 & 76.35 & 76.66 & 69.97 \\
    \midrule
    \multirow{4}{*}{Llama-3-70B-Instruct} & FP16 & 5.33  & 75.69 & 82.58 & 64.42 & 84.97 & 82.15 & 77.96 \\
    \cdashline{2-9} 
    & W4A8KV4-O1 & 5.82 & 77.11 & 82.42 & 66.72 & 84.47 & 81.72 & 78.49  \\
    & W4A8KV4-O2 & 5.96  & 77.74 & 81.97 & 65.87 & 84.93 & 81.56 & 78.41 \\
    & W4A4KV4-O1 & 6.49 & 75.77 & 81.51 & 65.96 & 83.8 & 81.28 & 77.66  \\
    & W4A4KV4-O2 & 6.80  & 75.93 & 80.64 & 64.76 & 83.88 & 81.23 & 77.29\\
    \midrule
    \end{tabular}
    \label{tab:other_models}
\end{table}

\begin{figure}[!h]
    \centering
    \begin{subfigure}[b]{0.45\textwidth}
    \centering
        \includegraphics[width=\textwidth]{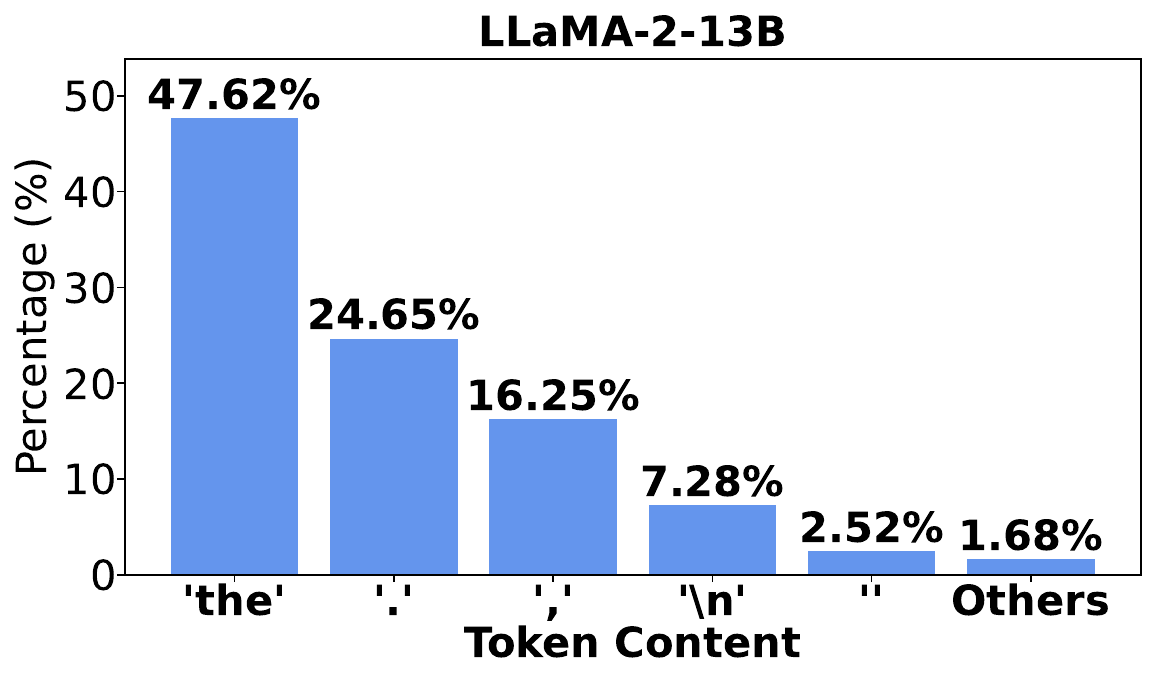}
        \vspace{-0.7em}
        \caption{Llama-2-13B}
        \label{fig:llama-2-13b-content}
    \end{subfigure}
    \hfill
    \begin{subfigure}[b]{0.45\textwidth}
    \centering
        \includegraphics[width=\textwidth]{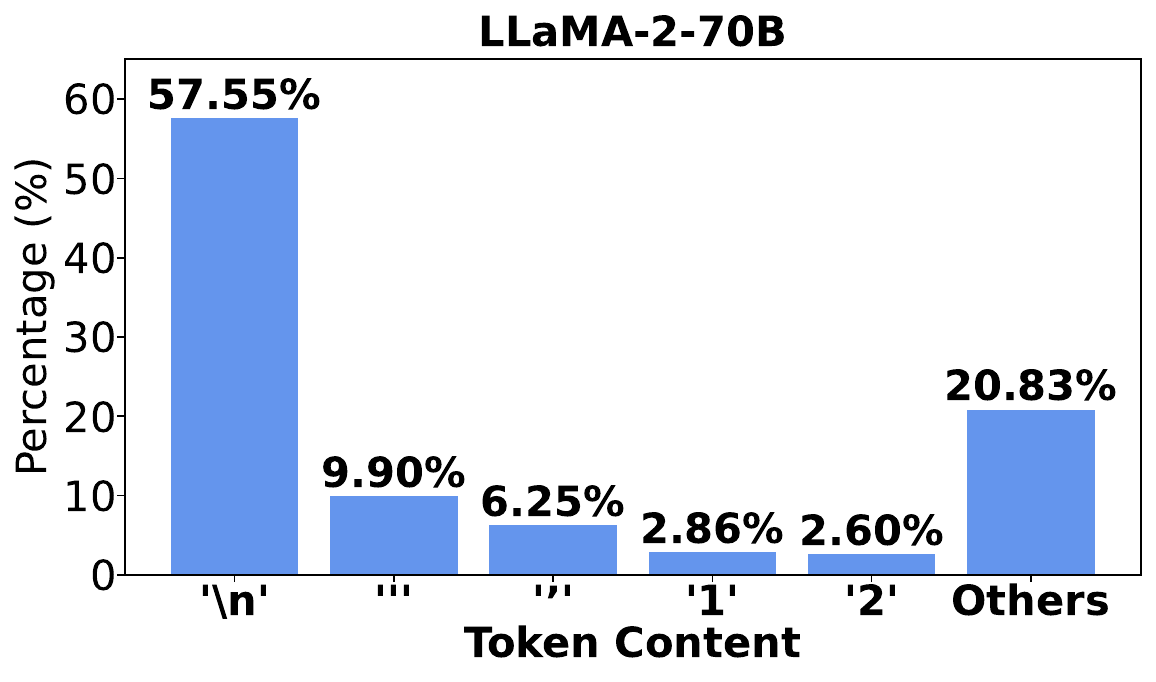}
        \vspace{-0.7em}
        \caption{Llama-2-70B}
        \label{fig:llama-2-70b-content}
    \end{subfigure}
    \hfill
    \begin{subfigure}[b]{0.45\textwidth}
    \centering
        \includegraphics[width=\textwidth]{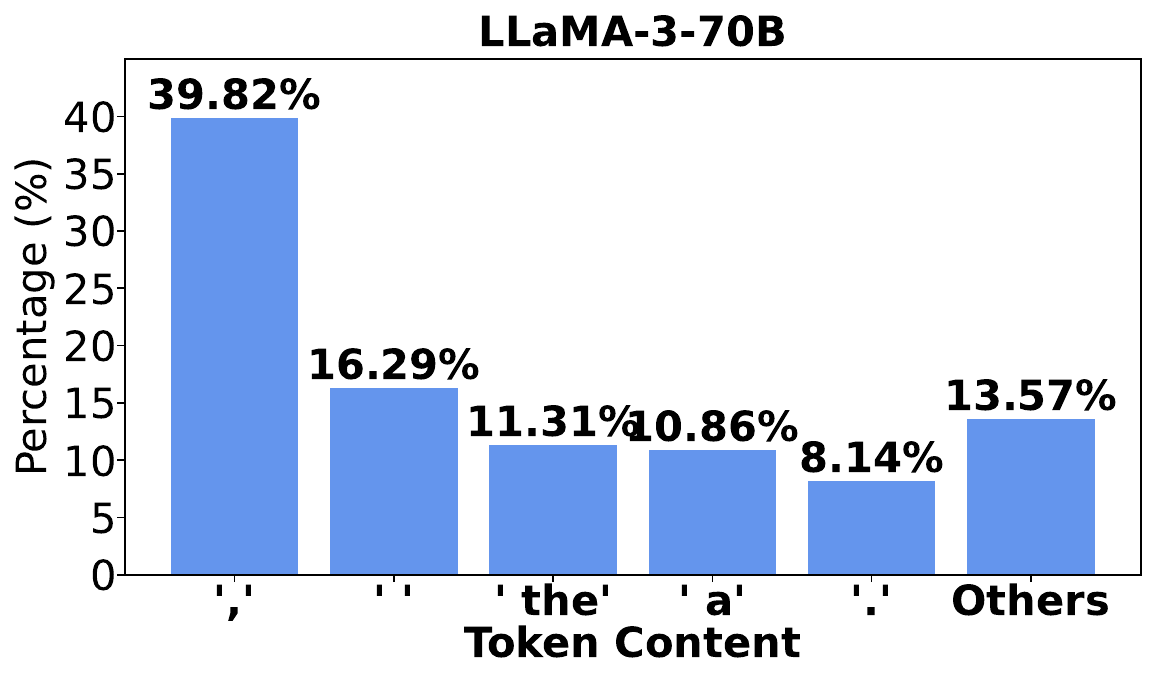}
        \vspace{-0.7em}
        \caption{Llama-3-70B}
        \label{fig:llama-3-70b-content}
    \end{subfigure}
    \hfill
    \begin{subfigure}[b]{0.45\textwidth}
    \centering
        \includegraphics[width=\textwidth]{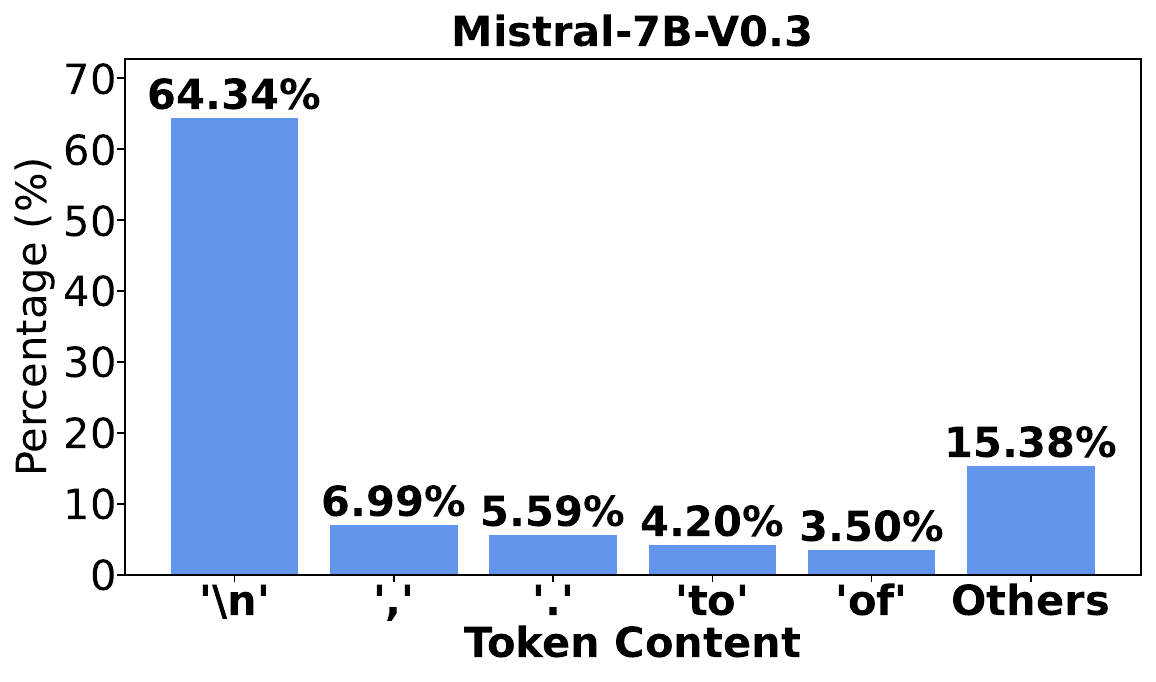}
        \vspace{-0.7em}
        \caption{Mistral-7B-v0.3}
        \label{fig:mistral-7b-v0.3}
    \end{subfigure}
    \hfill
    \caption{\textbf{Content of outlier tokens in different models.} Note that we do not count the outlier tokens situated at the initial token.}
    \label{fig:other-models-contents}
\end{figure}

\section{More Visualizations}\label{sec:more_visualization}

\subsection{Outlier Token}
In Figure~\ref{fig:other-models-contents}, we showcase the four most frequently occurring outlier tokens in Llama-2-\{13B,70B\}, Llama-3-70B, and Mistral-7B-v0.3. 
Specifically, Table~\ref{tab:prefixed_token} selects the top-$o$ high-frequent outlier tokens as the prefixed tokens.
It is important to note that we do not visualize the outlier tokens in Llama-3-8B and Qwen-2-7B because all the outlier tokens in these two models appear in the initial tokens.

\subsection{Magnitude Distribution}
\begin{figure}[htbp]
    \centering
    \begin{subfigure}[b]{0.95\textwidth}
    \centering
        \includegraphics[width=0.9\textwidth]{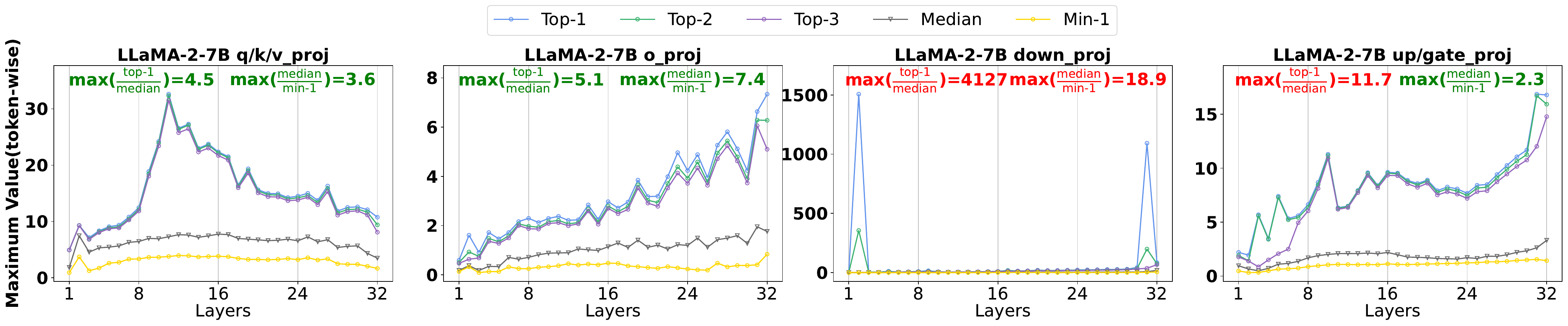}
        \vspace{-0.7em}
        \caption{Original distribution}
        \label{fig:llama-2-7b-input}
    \end{subfigure}
    \hfill
    \begin{subfigure}[b]{0.95\textwidth}
    \centering
        \includegraphics[width=0.9\textwidth]{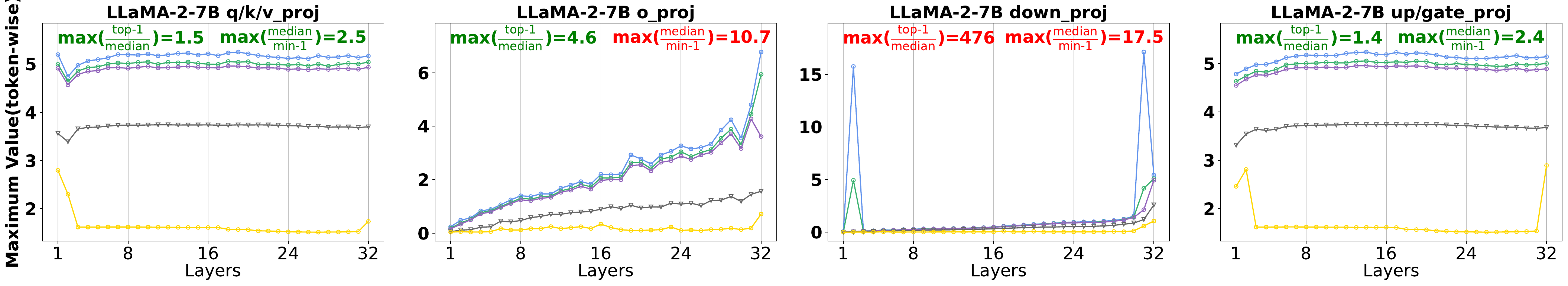}
        \vspace{-0.7em}
        \caption{Rotation}
        \label{fig:llama-2-7b-input-rotate}
    \end{subfigure}
    \hfill
    \begin{subfigure}[b]{0.95\textwidth}
    \centering
        \includegraphics[width=0.9\textwidth]{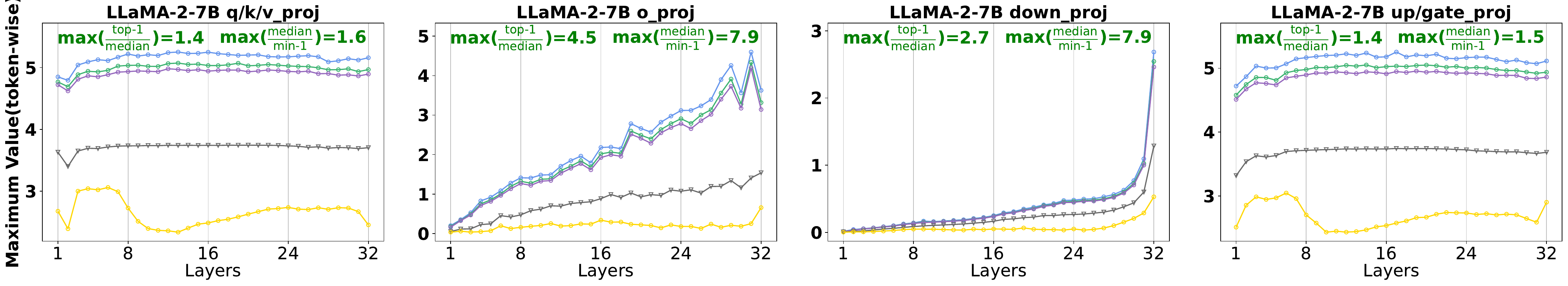}
        \vspace{-0.7em}
        \caption{\methodshort (ours)}
        \label{fig:llama-2-7b-input-rotate-prefix}
    \end{subfigure}
    \caption{\textbf{Distribution of token-wise maximum values for linear layers inputs in Llama-2-7B.} Top-$N$ indicates the $N$-th largest value, Min-$N$ indicates the $N$-th smallest value.}
    \label{fig:llama-2-7b-input-all}
\end{figure}

\begin{figure}[!t]
    \centering
    \begin{subfigure}[b]{0.95\textwidth}
    \centering
        \includegraphics[width=0.75\textwidth]{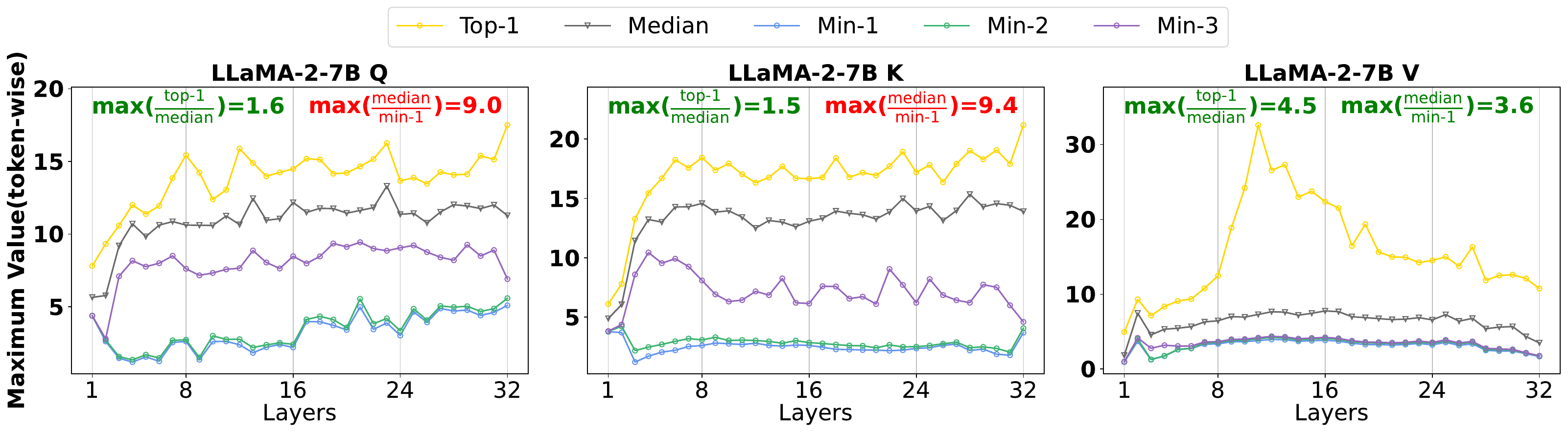}
        \vspace{-0.7em}
        \caption{Original distribution}
        \label{fig:llama-2-7b-output}
    \end{subfigure}
    \hfill
    \begin{subfigure}[b]{0.95\textwidth}
    \centering
        \includegraphics[width=0.75\textwidth]{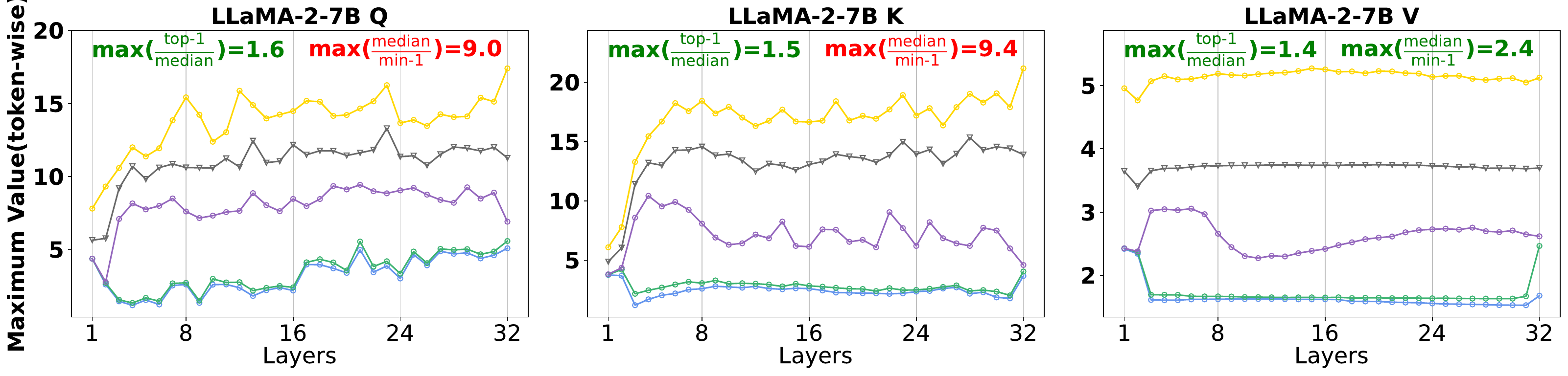}
        \vspace{-0.7em}
        \caption{Rotation}
        \label{fig:llama-2-7b-output-rotate}
    \end{subfigure}
    \hfill
    \begin{subfigure}[b]{0.95\textwidth}
    \centering
        \includegraphics[width=0.75\textwidth]{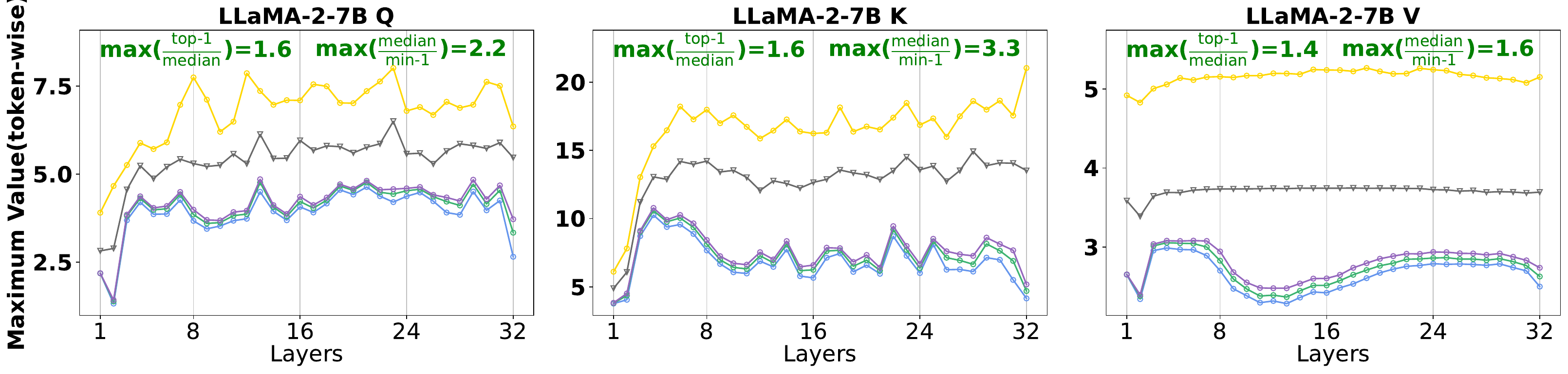}
        \vspace{-0.7em}
        \caption{\methodshort (ours)}
        \label{fig:llama-2-7b-output-rotate-prefix}
    \end{subfigure}
    \caption{\textbf{Distribution of token-wise maximum values for $\mathbf{Q}$/$\mathbf{K}$/$\mathbf{V}$ in Llama-2-7B.} Same present rules as Figure ~\ref{fig:llama-2-7b-input} except that ratios greater than 5 are marked with red.}
    \label{fig:llama-2-7b-output-all}
\end{figure}

\begin{figure}[!ht]
    \centering
    \begin{subfigure}[b]{0.95\textwidth}
    \centering
        \includegraphics[width=0.9\textwidth]{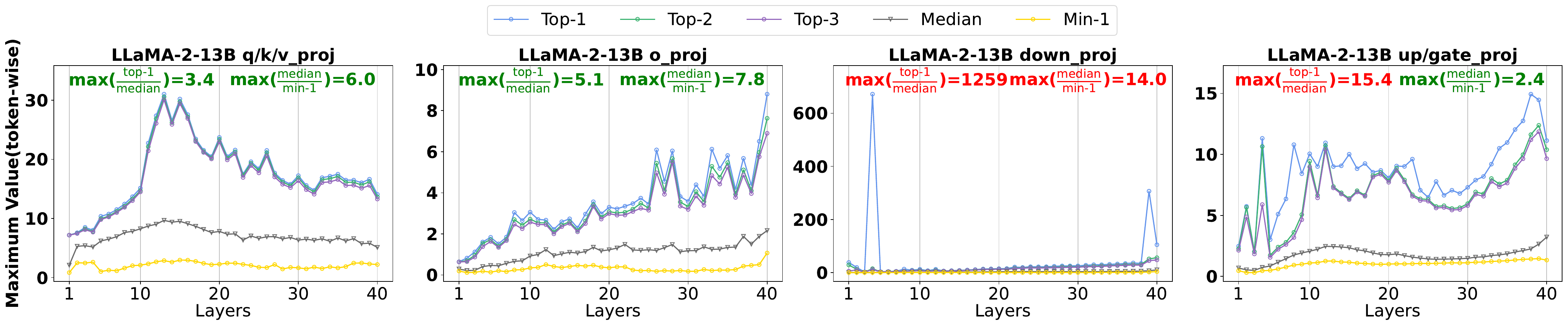}
        \vspace{-0.7em}
        \caption{Original distribution}
        \label{fig:llama-2-13b-input}
    \end{subfigure}
    \hfill
    \begin{subfigure}[b]{0.95\textwidth}
    \centering
        \includegraphics[width=0.9\textwidth]{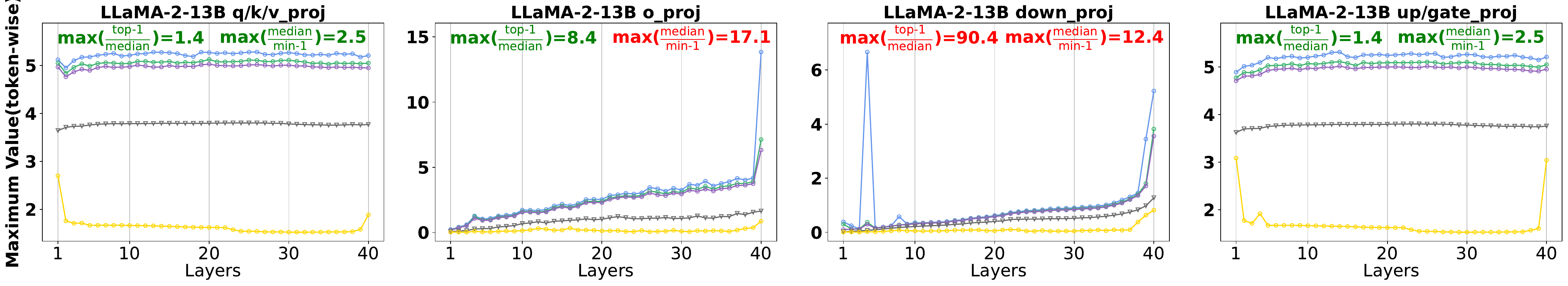}
        \vspace{-0.7em}
        \caption{Rotation}
        \label{fig:llama-2-13b-input-rotate}
    \end{subfigure}
    \hfill
    \begin{subfigure}[b]{0.95\textwidth}
    \centering
        \includegraphics[width=0.9\textwidth]{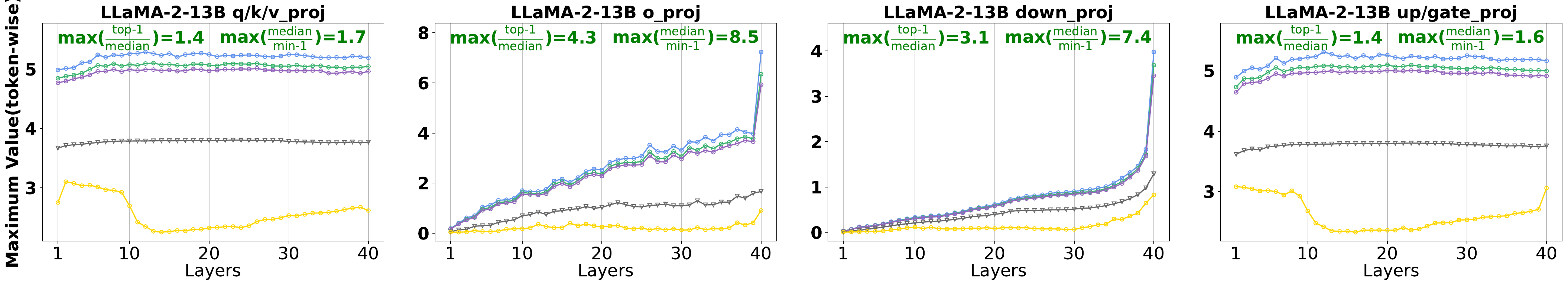}
        \vspace{-0.7em}
        \caption{\methodshort (ours)}
        \label{fig:llama-2-13b-input-rotate-prefix}
    \end{subfigure}
    \caption{\textbf{Distribution of token-wise maximum values for linear layers inputs in Llama-2-13b.}}
    \label{fig:llama-2-13b-input-all}
\end{figure}

\begin{figure}[!h]
    \centering
    \begin{subfigure}[b]{0.95\textwidth}
    \centering
        \includegraphics[width=0.75\textwidth]{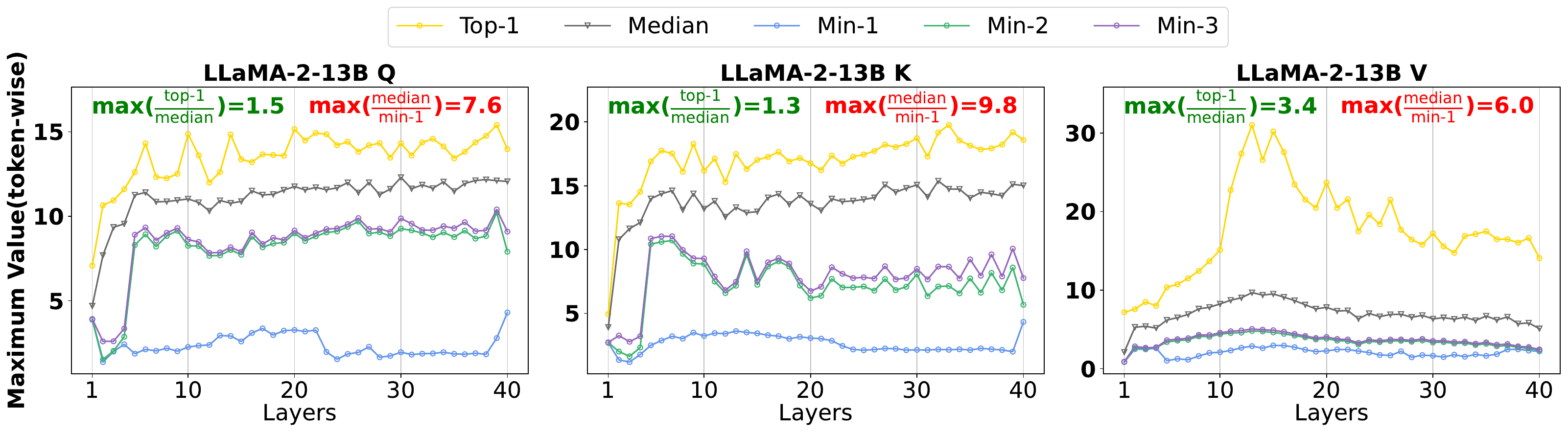}
        \vspace{-0.7em}
        \caption{Original distribution}
        \label{fig:llama-2-13b-output}
    \end{subfigure}
    \hfill
    \begin{subfigure}[b]{0.95\textwidth}
    \centering
        \includegraphics[width=0.75\textwidth]{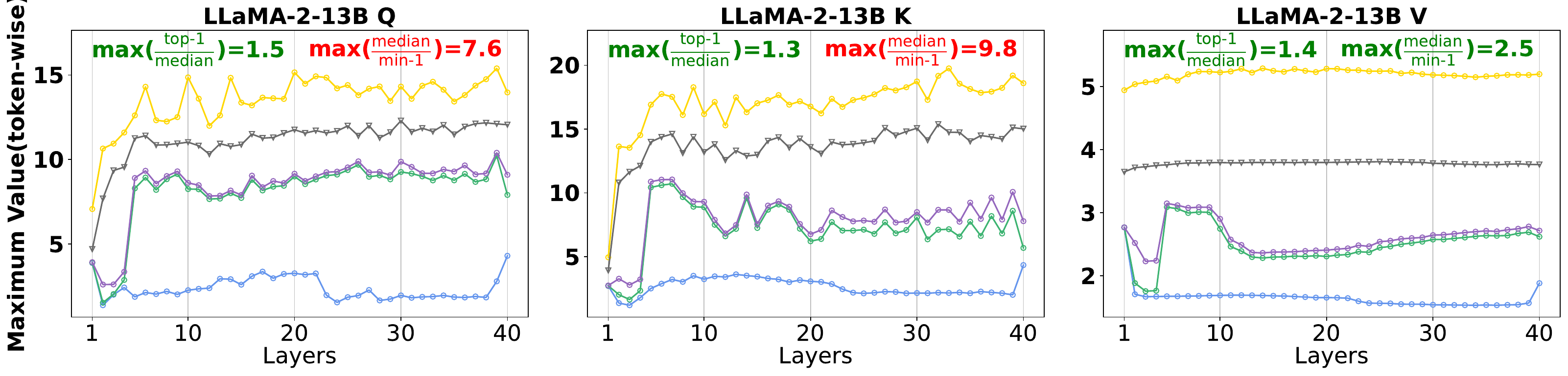}
        \vspace{-0.7em}
        \caption{Rotation}
        \label{fig:llama-2-13b-output-rotate}
    \end{subfigure}
    \hfill
    \begin{subfigure}[b]{0.95\textwidth}
    \centering
        \includegraphics[width=0.75\textwidth]{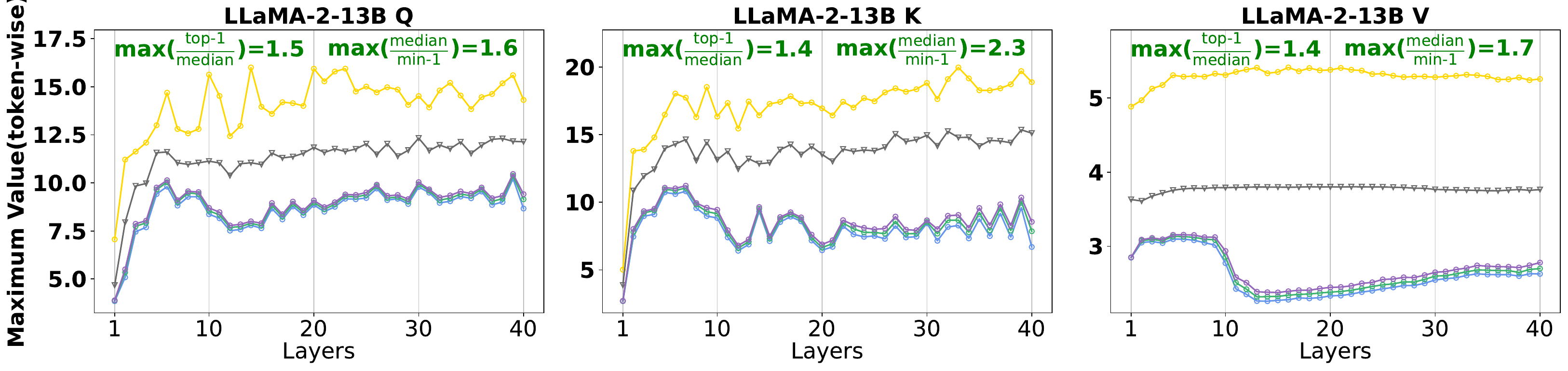}
        \vspace{-0.7em}
        \caption{\methodshort (ours)}
        \label{fig:llama-2-13b-output-rotate-prefix}
    \end{subfigure}
    \caption{\textbf{Distribution of token-wise maximum values for $\mathbf{Q}$/$\mathbf{K}$/$\mathbf{V}$ in Llama-2-13b.} Same present rules as Figure ~\ref{fig:llama-2-13b-input} except that ratios greater than 5 are marked with red.}
    \label{fig:llama-2-13b-output-all}
\end{figure}

\begin{figure}[!t]
    \centering
    \begin{subfigure}[b]{0.95\textwidth}
    \centering
        \includegraphics[width=0.9\textwidth]{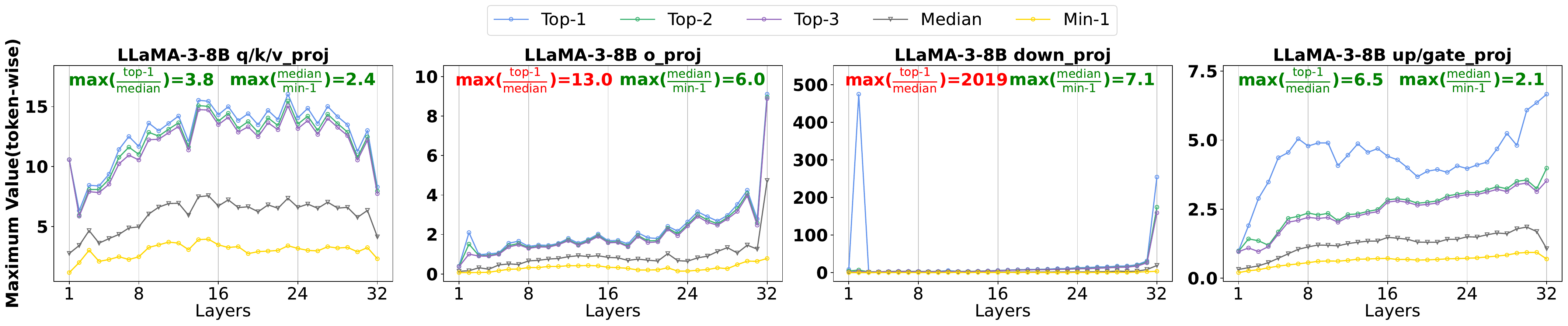}
        \vspace{-0.7em}
        \caption{Original distribution}
        \label{fig:llama-3-8b-input}
    \end{subfigure}
    \hfill
    \begin{subfigure}[b]{0.95\textwidth}
    \centering
        \includegraphics[width=0.9\textwidth]{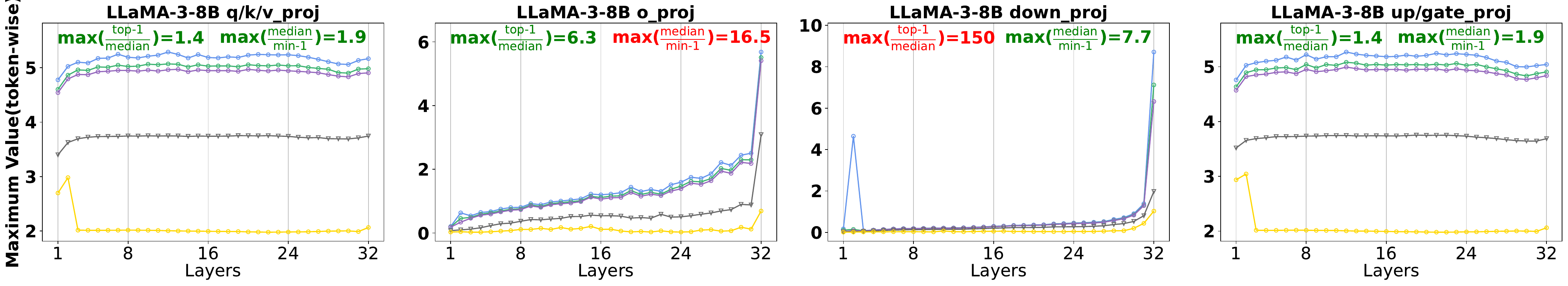}
        \vspace{-0.7em}
        \caption{Rotation}
        \label{fig:llama-3-8b-input-rotate}
    \end{subfigure}
    \hfill
    \begin{subfigure}[b]{0.95\textwidth}
    \centering
        \includegraphics[width=0.9\textwidth]{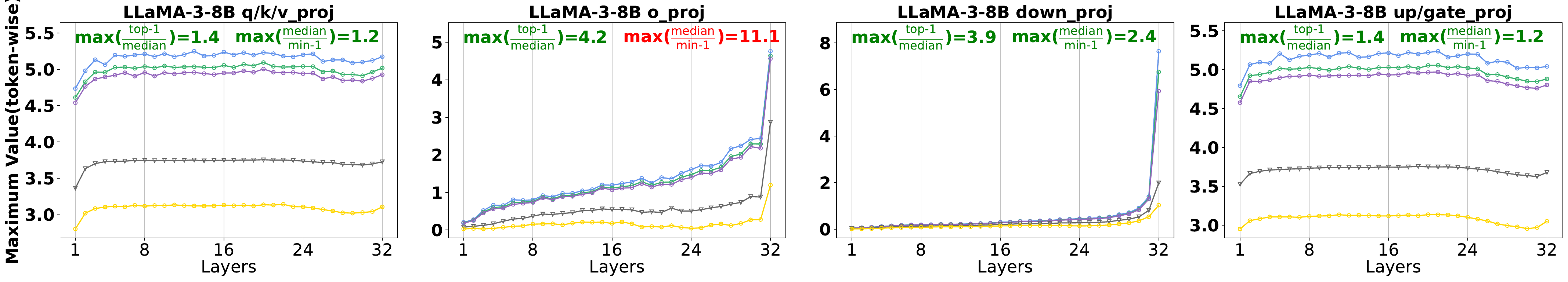}
        \vspace{-0.7em}
        \caption{\methodshort (ours)}
        \label{fig:llama-3-8b-input-rotate-prefix}
    \end{subfigure}
    \caption{\textbf{Distribution of token-wise maximum values for linear layers inputs in Llama-3-8b.}}
    \label{fig:llama-3-8b-input-all}
\end{figure}

\begin{figure}[!ht]
    \centering
    \begin{subfigure}[b]{0.95\textwidth}
    \centering
        \includegraphics[width=0.75\textwidth]{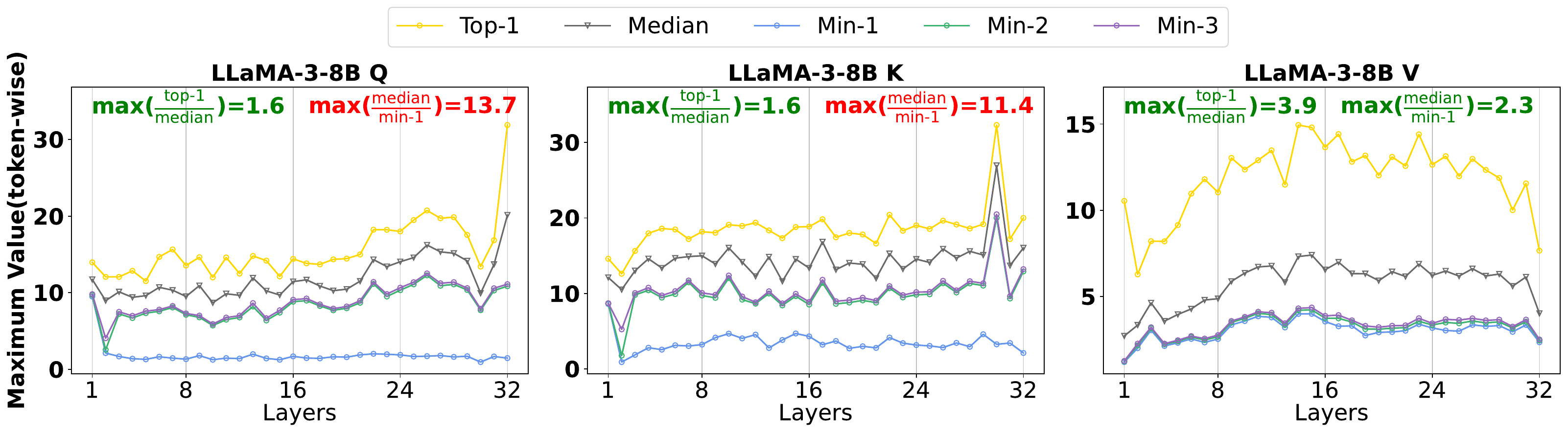}
        \vspace{-0.7em}
        \caption{Original distribution}
        \label{fig:llama-3-8b-output}
    \end{subfigure}
    \hfill
    \begin{subfigure}[b]{0.95\textwidth}
    \centering
        \includegraphics[width=0.75\textwidth]{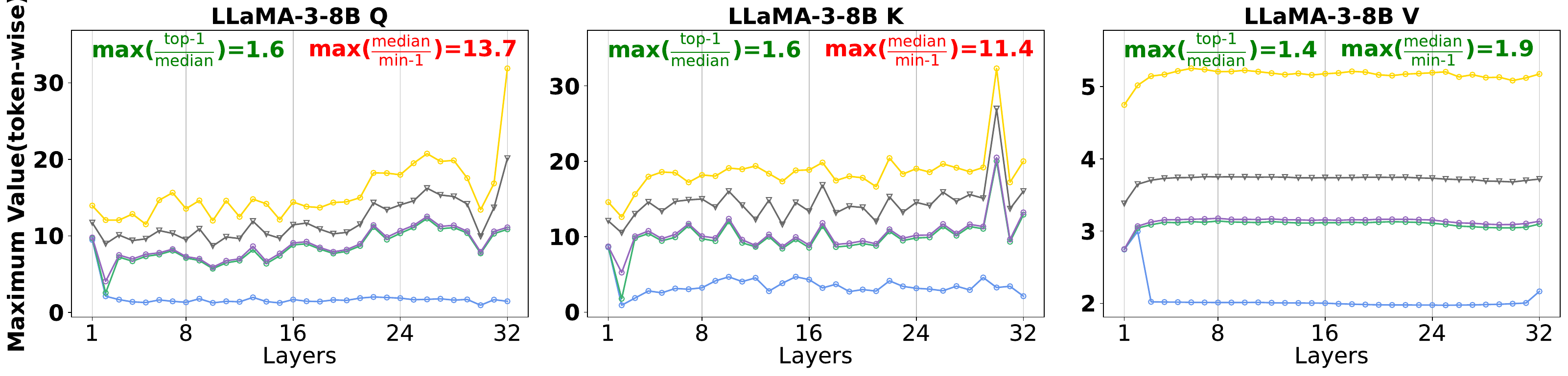}
        \vspace{-0.7em}
        \caption{Rotation}
        \label{fig:llama-3-8b-output-rotate}
    \end{subfigure}
    \hfill
    \begin{subfigure}[b]{0.95\textwidth}
    \centering
        \includegraphics[width=0.75\textwidth]{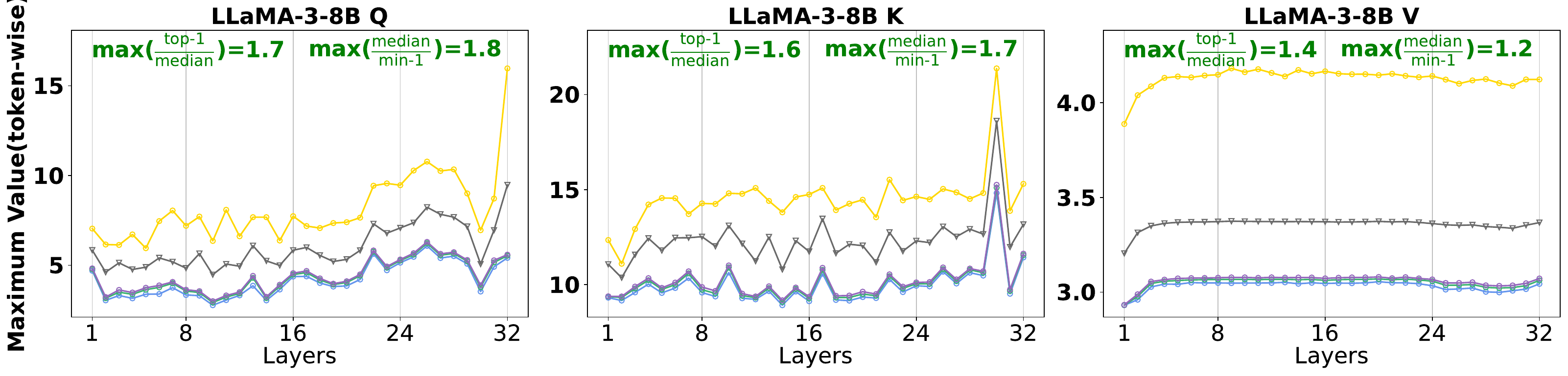}
        \vspace{-0.7em}
        \caption{\methodshort (ours)}
        \label{fig:llama-3-8b-output-rotate-prefix}
    \end{subfigure}
    \caption{\textbf{Distribution of token-wise maximum values for $\mathbf{Q}$/$\mathbf{K}$/$\mathbf{V}$ in Llama-3-8B.}}
    \label{fig:llama-3-8b-output-all}
\end{figure}

\begin{figure}[!ht]
    \centering
    \begin{subfigure}[b]{0.95\textwidth}
    \centering
        \includegraphics[width=0.9\textwidth]{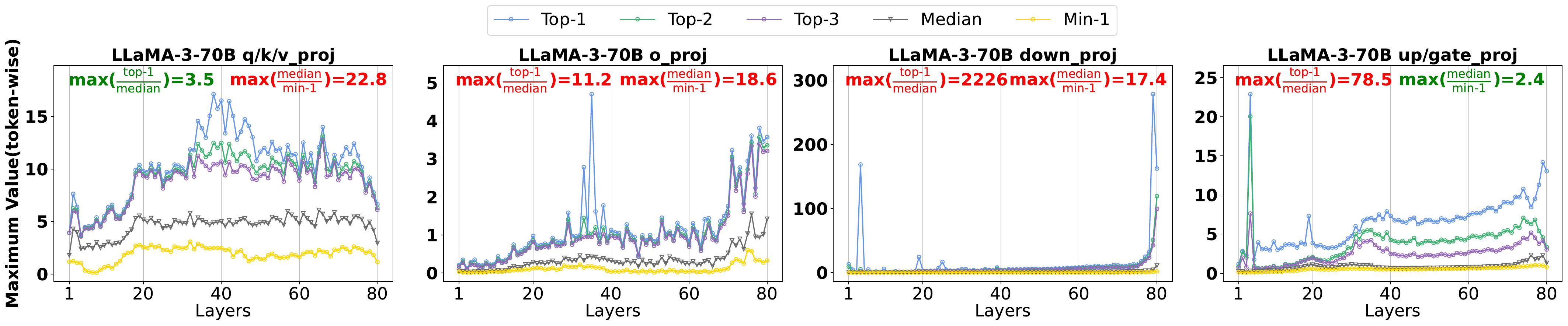}
        \vspace{-0.7em}
        \caption{Original distribution}
        \label{fig:llama-3-70b-input}
    \end{subfigure}
    \hfill
    \begin{subfigure}[b]{0.95\textwidth}
    \centering
        \includegraphics[width=0.9\textwidth]{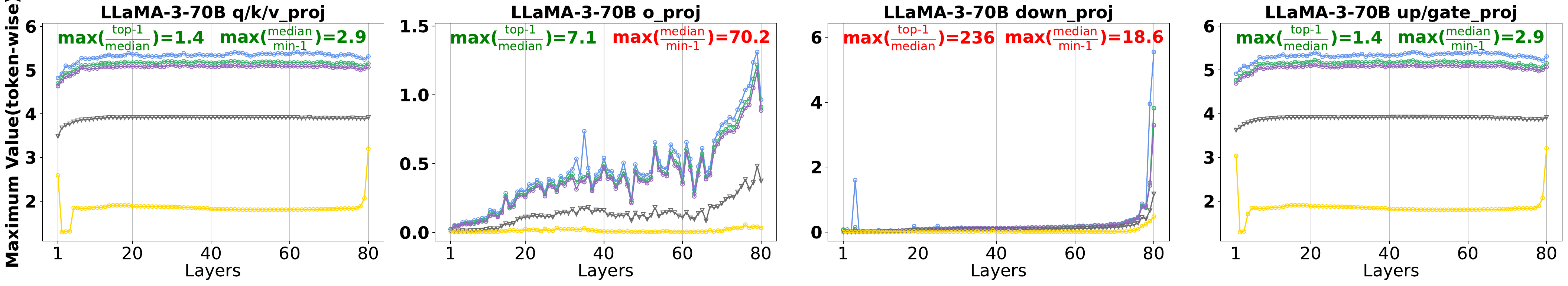}
        \vspace{-0.7em}
        \caption{Rotation}
        \label{fig:llama-3-70b-input-rotate}
    \end{subfigure}
    \hfill
    \begin{subfigure}[b]{0.95\textwidth}
    \centering
        \includegraphics[width=0.9\textwidth]{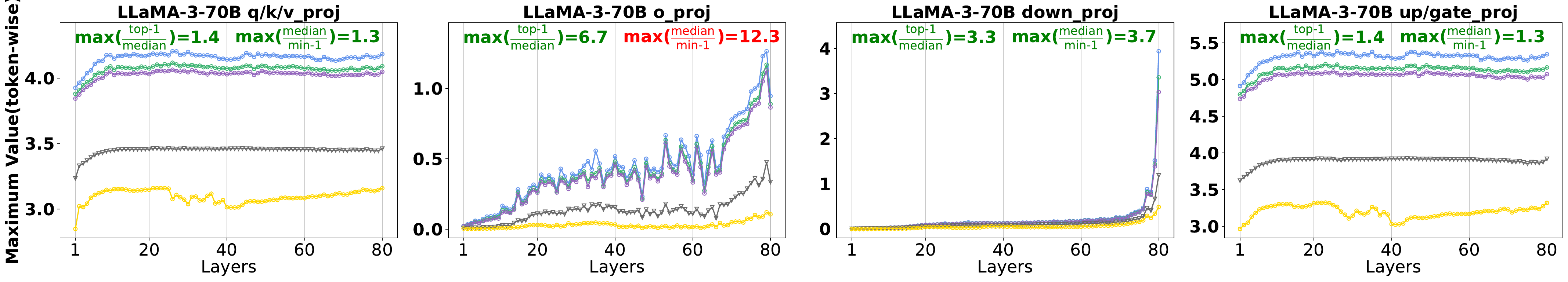}
        \vspace{-0.7em}
        \caption{\methodshort (ours)}
        \label{fig:llama-3-70b-input-rotate-prefix}
    \end{subfigure}
    \caption{\textbf{Distribution of token-wise maximum values for linear layers inputs in Llama-3-70B.}}
    \label{fig:llama-3-70b-input-all}
\end{figure}

\begin{figure}[!ht]
    \centering
    \begin{subfigure}[b]{0.95\textwidth}
    \centering
        \includegraphics[width=0.75\textwidth]{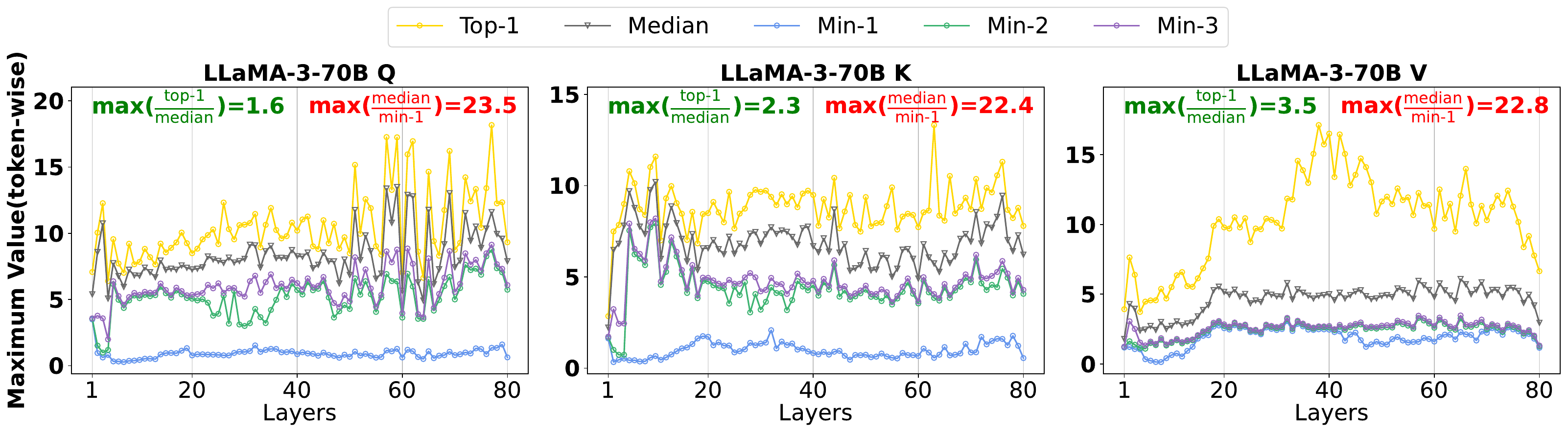}
        \vspace{-0.7em}
        \caption{Original distribution}
        \label{fig:llama-3-70b-output}
    \end{subfigure}
    \hfill
    \begin{subfigure}[b]{0.95\textwidth}
    \centering
        \includegraphics[width=0.75\textwidth]{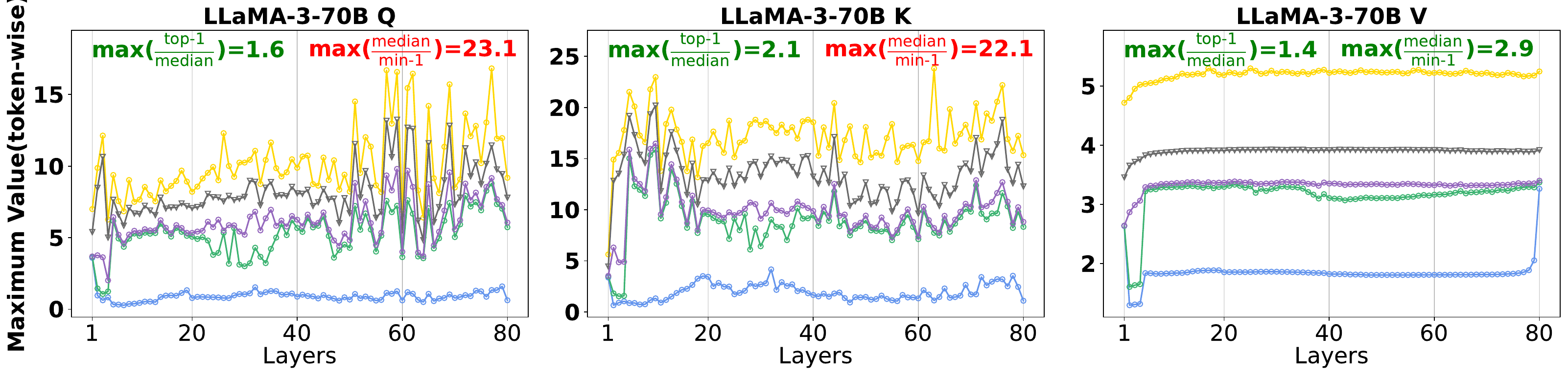}
        \vspace{-0.7em}
        \caption{Rotation}
        \label{fig:llama-3-70b-output-rotate}
    \end{subfigure}
    \hfill
    \begin{subfigure}[b]{0.95\textwidth}
    \centering
        \includegraphics[width=0.75\textwidth]{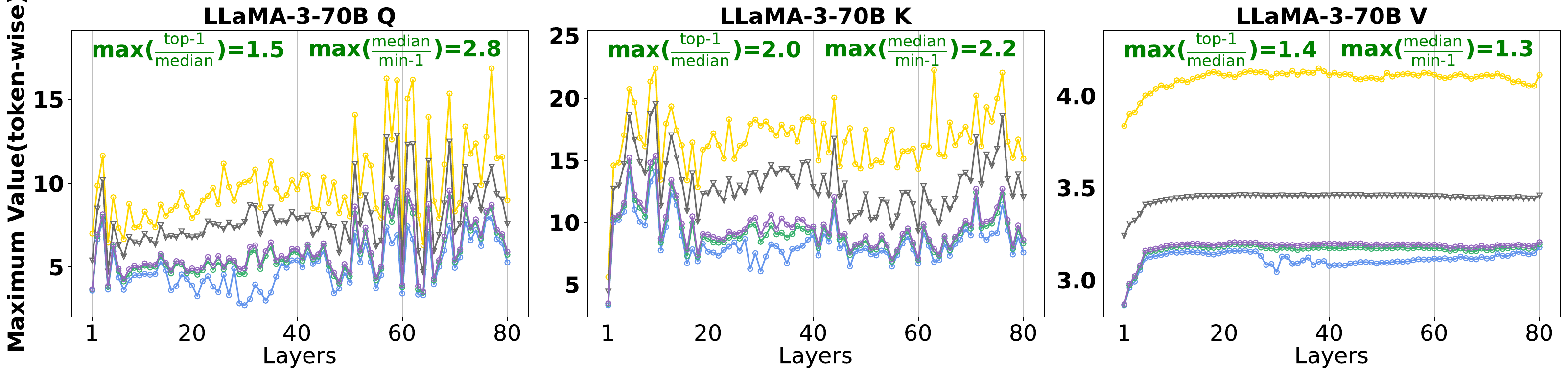}
        \vspace{-0.7em}
        \caption{\methodshort (ours)}
        \label{fig:llama-3-70b-output-rotate-prefix}
    \end{subfigure}
    \caption{\textbf{Distribution of token-wise maximum values for $\mathbf{Q}$/$\mathbf{K}$/$\mathbf{V}$ in Llama-3-70B.}}
    \label{fig:llama-3-70b-output-all}
\end{figure}

\begin{figure}[!ht]
    \centering
    \begin{subfigure}[b]{0.95\textwidth}
    \centering
        \includegraphics[width=0.9\textwidth]{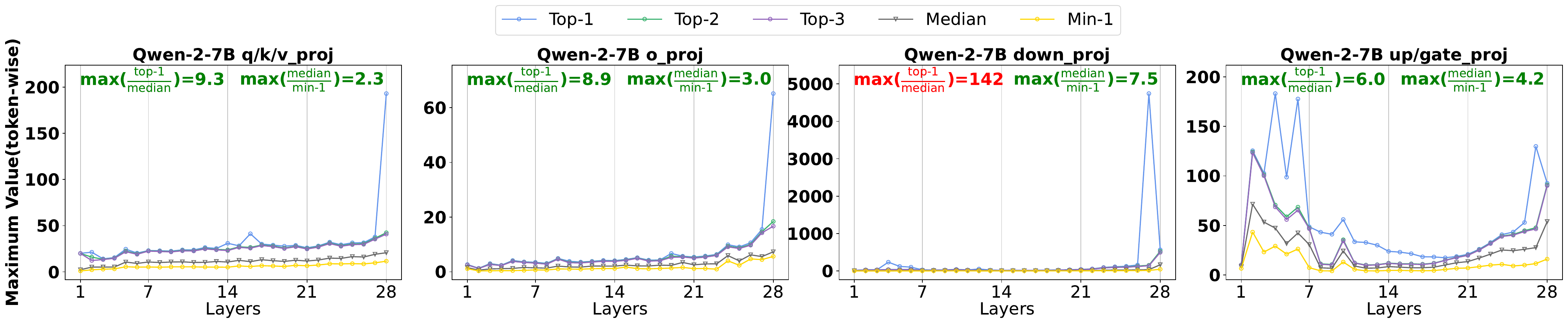}
        \vspace{-0.7em}
        \caption{Original distribution}
        \label{fig:qwen-2-7b-input}
    \end{subfigure}
    \hfill
    \begin{subfigure}[b]{0.95\textwidth}
    \centering
        \includegraphics[width=0.9\textwidth]{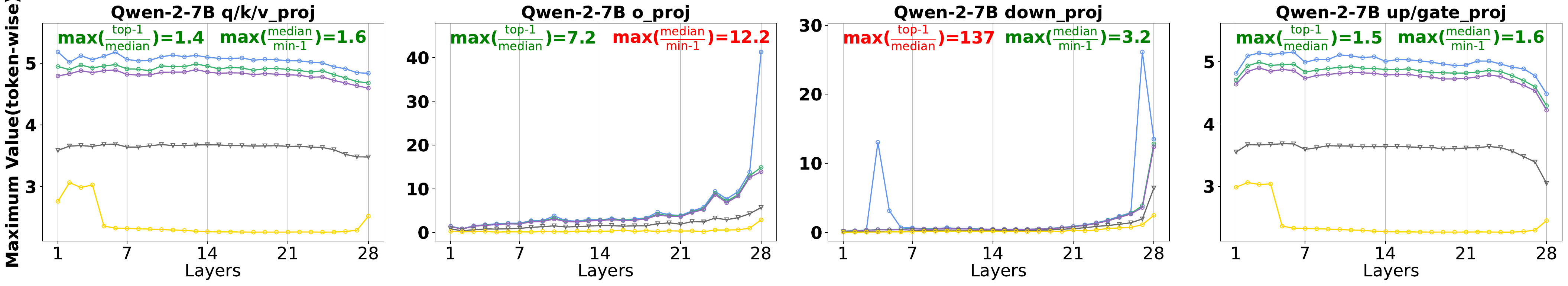}
        \vspace{-0.7em}
        \caption{Rotation}
        \label{fig:qwen-2-7b-input-rotate}
    \end{subfigure}
    \hfill
    \begin{subfigure}[b]{0.95\textwidth}
    \centering
        \includegraphics[width=0.9\textwidth]{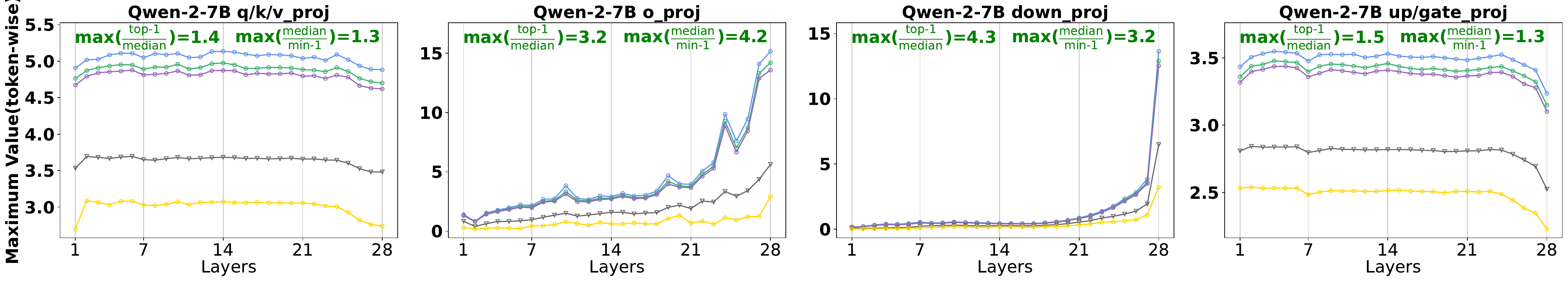}
        \vspace{-0.7em}
        \caption{\methodshort (ours)}
        \label{fig:qwen-2-7b-input-rotate-prefix}
    \end{subfigure}
    \caption{\textbf{Distribution of token-wise maximum values for linear layers inputs in Qwen-2-7B.}}
    \label{fig:qwen-2-7b-input-all}
\end{figure}

\begin{figure}[!ht]
    \centering
    \begin{subfigure}[b]{0.95\textwidth}
    \centering
        \includegraphics[width=0.75\textwidth]{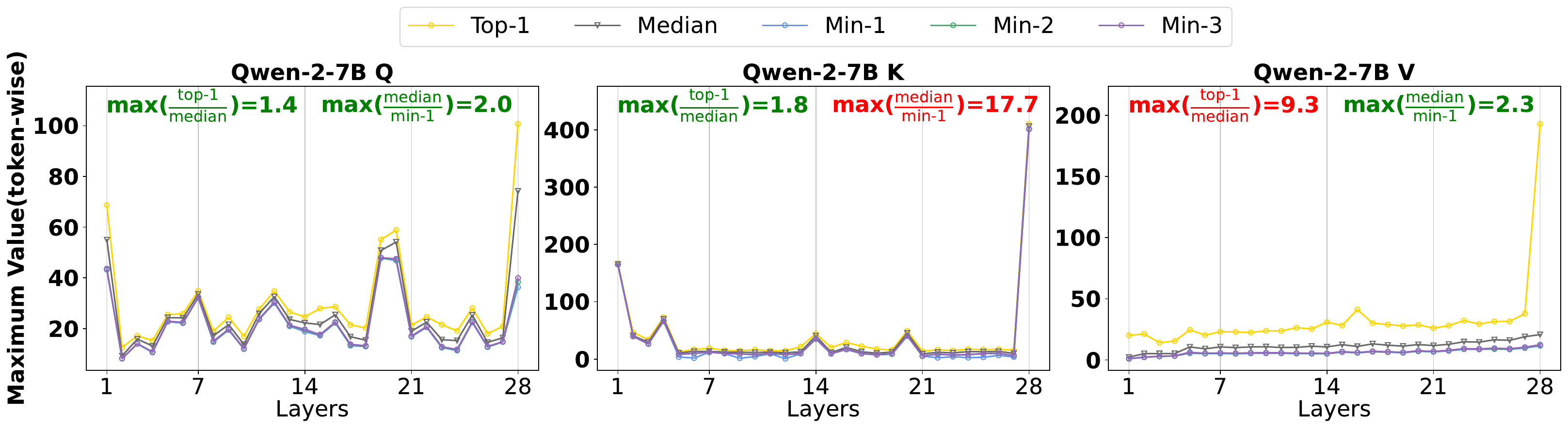}
        \vspace{-0.7em}
        \caption{Original distribution}
        \label{fig:qwen-2-7b-output}
    \end{subfigure}
    \hfill
    \begin{subfigure}[b]{0.95\textwidth}
    \centering
        \includegraphics[width=0.75\textwidth]{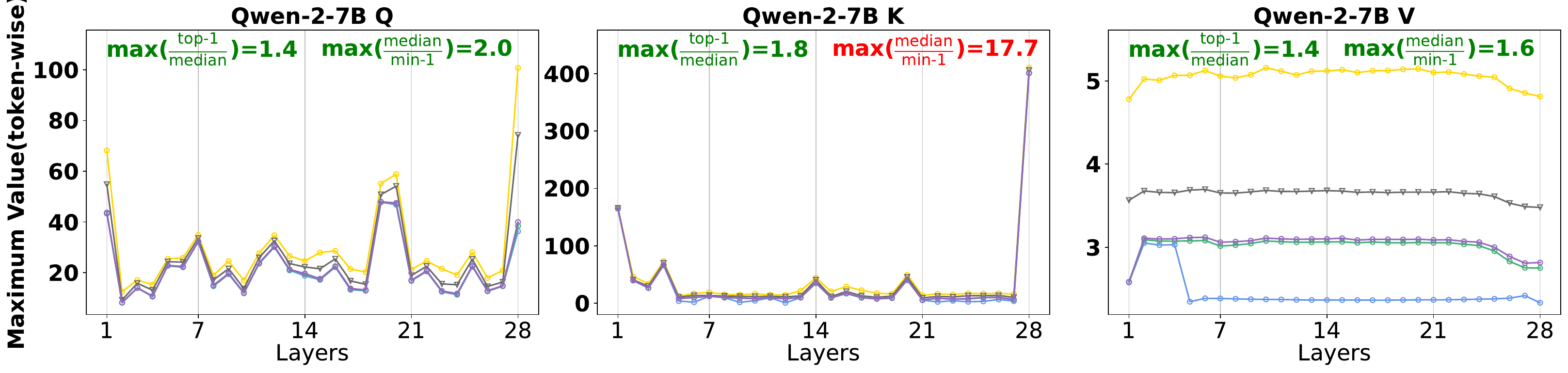}
        \vspace{-0.7em}
        \caption{Rotation}
        \label{fig:qwen-2-7b-output-rotate}
    \end{subfigure}
    \hfill
    \begin{subfigure}[b]{0.95\textwidth}
    \centering
        \includegraphics[width=0.75\textwidth]{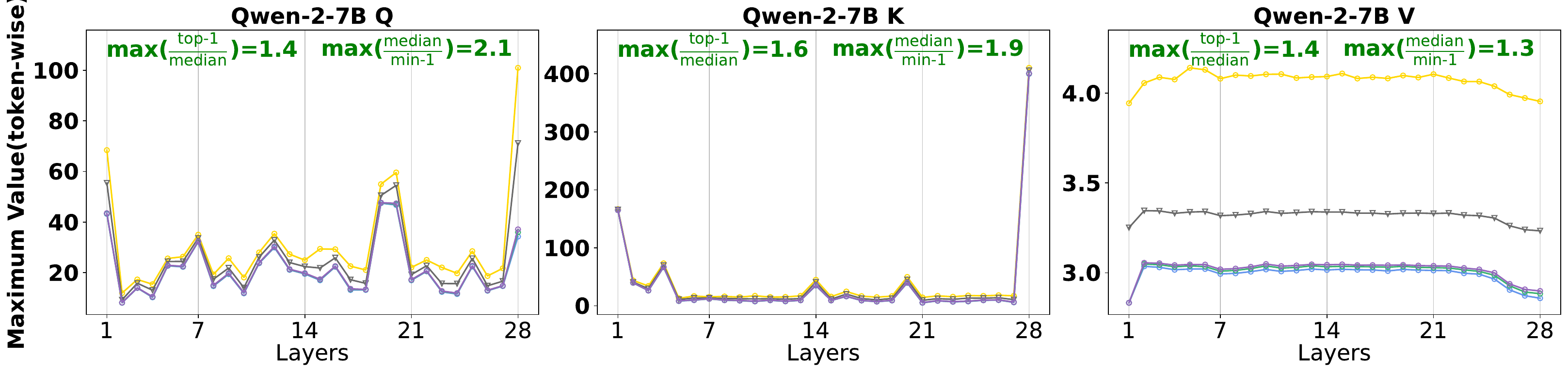}
        \vspace{-0.7em}
        \caption{\methodshort (ours)}
        \label{fig:qwen-2-7b-output-rotate-prefix}
    \end{subfigure}
    \caption{\textbf{Distribution of token-wise maximum values for $\mathbf{Q}$/$\mathbf{K}$/$\mathbf{V}$ in Qwen-2-7B.}}
    \label{fig:qwen-2-7b-output-all}
\end{figure}

\begin{figure}[!ht]
    \centering
    \begin{subfigure}[b]{0.95\textwidth}
    \centering
        \includegraphics[width=0.9\textwidth]{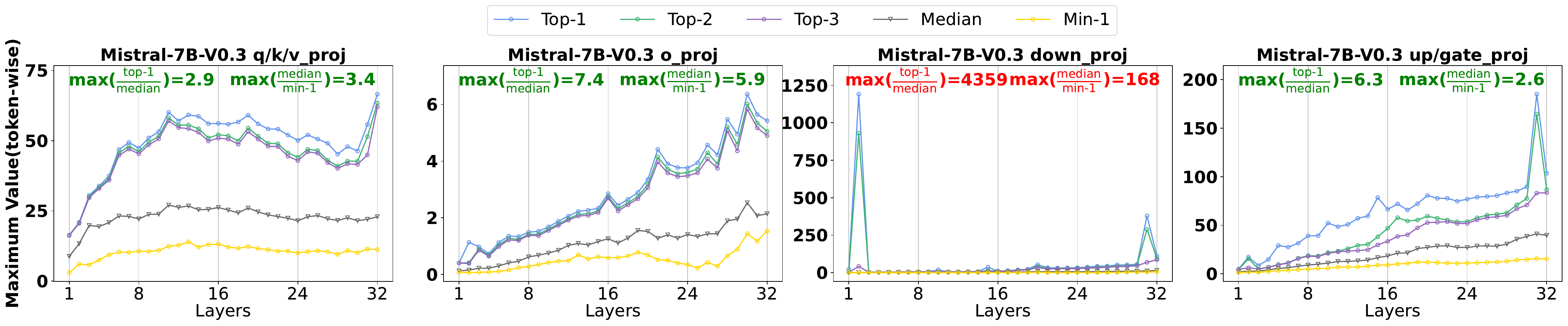}
        \vspace{-0.7em}
        \caption{Original distribution}
        \label{fig:mistral-7b-v0.3-input}
    \end{subfigure}
    \hfill
    \begin{subfigure}[b]{0.95\textwidth}
    \centering
        \includegraphics[width=0.9\textwidth]{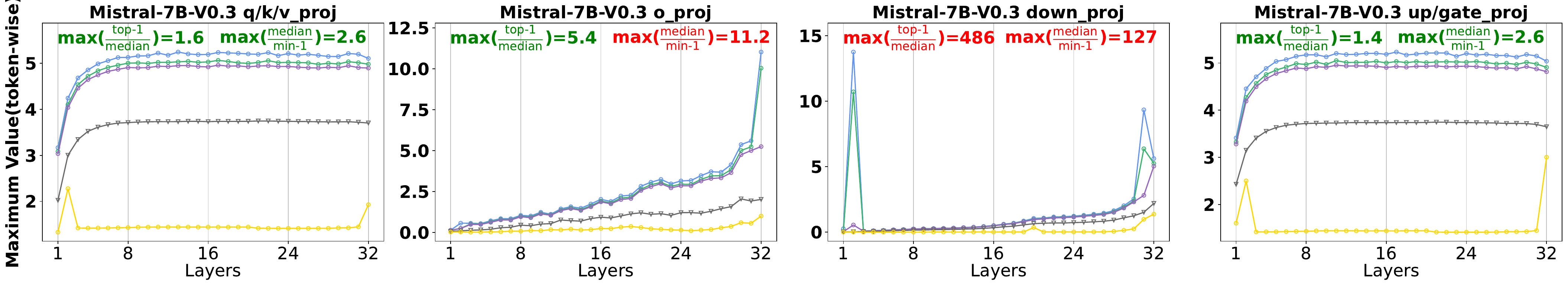}
        \vspace{-0.7em}
        \caption{Rotation}
        \label{fig:mistral-7b-v0.3-input-rotate}
    \end{subfigure}
    \hfill
    \begin{subfigure}[b]{0.95\textwidth}
    \centering
        \includegraphics[width=0.9\textwidth]{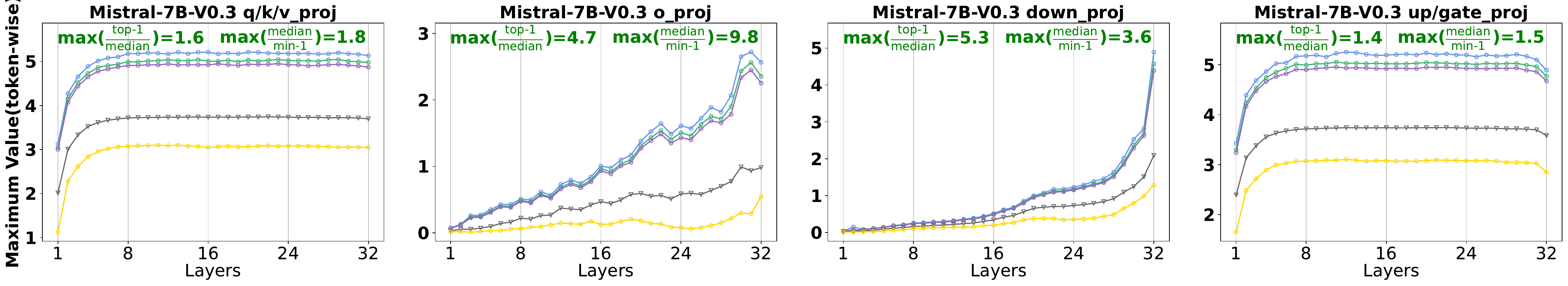}
        \vspace{-0.7em}
        \caption{\methodshort (ours)}
        \label{fig:mistral-7b-v0.3-input-rotate-prefix}
    \end{subfigure}
    \caption{\textbf{Distribution of token-wise maximum values for linear layers inputs in Mistral-7B-v0.3.}}
    \label{fig:mistral-7b-v0.3-input-all}
\end{figure}

\begin{figure}[!ht]
    \centering
    \begin{subfigure}[b]{0.95\textwidth}
    \centering
        \includegraphics[width=0.75\textwidth]{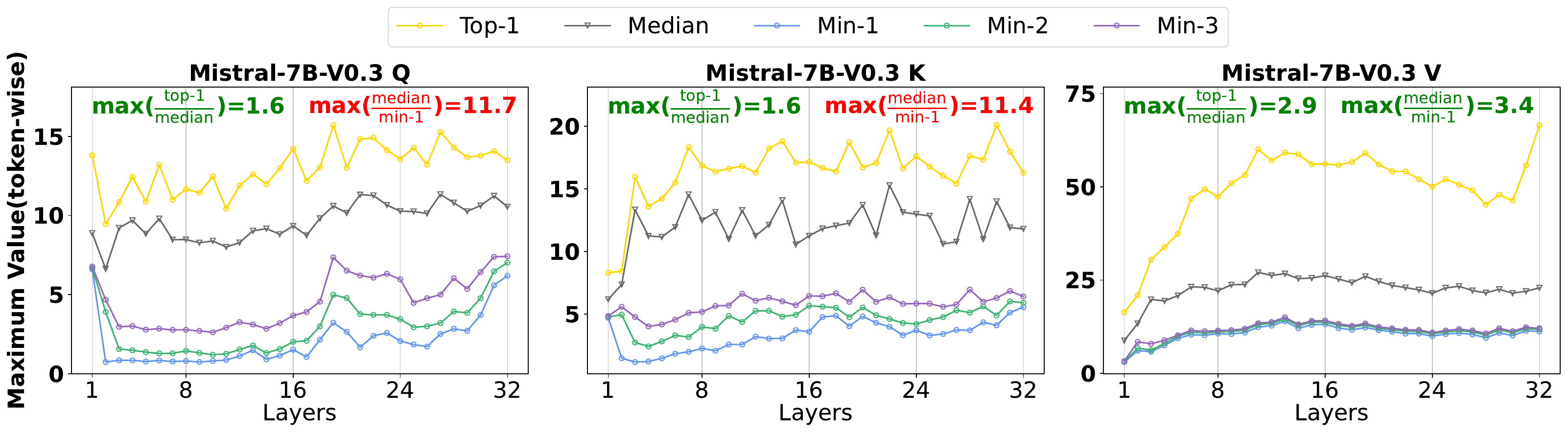}
        \vspace{-0.7em}
        \caption{Original distribution}
        \label{fig:mistral-7b-v0.3-output}
    \end{subfigure}
    \hfill
    \begin{subfigure}[b]{0.95\textwidth}
    \centering
        \includegraphics[width=0.75\textwidth]{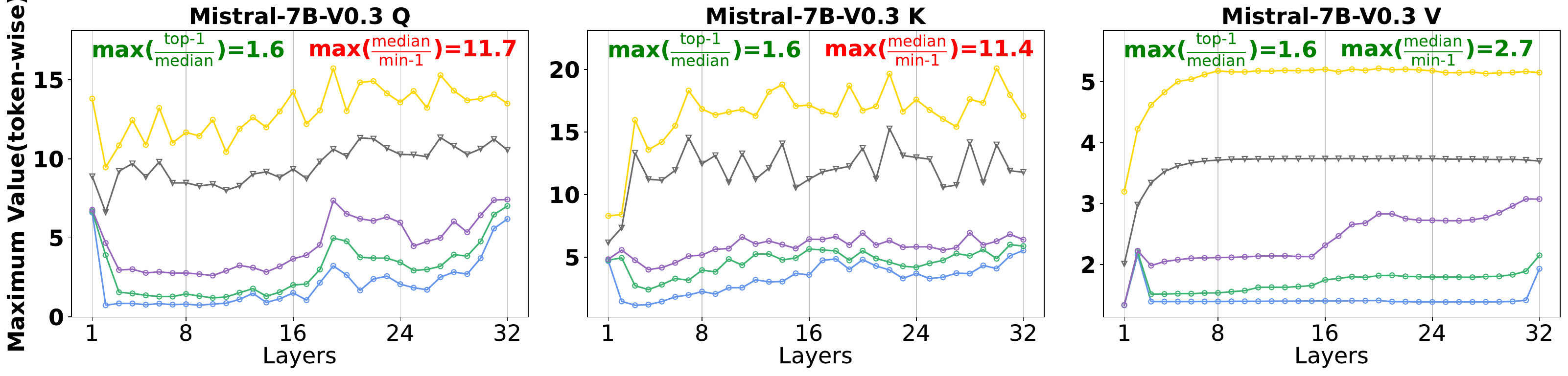}
        \vspace{-0.7em}
        \caption{Rotation}
        \label{fig:mistral-7b-v0.3-output-rotate}
    \end{subfigure}
    \hfill
    \begin{subfigure}[b]{0.95\textwidth}
    \centering
        \includegraphics[width=0.75\textwidth]{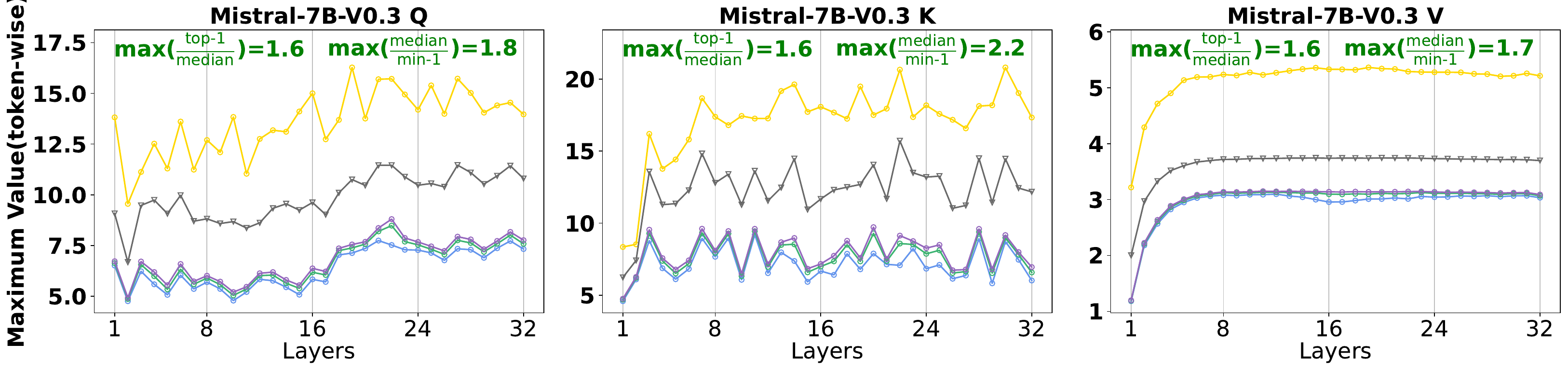}
        \vspace{-0.7em}
        \caption{\methodshort (ours)}
        \label{fig:mistral-7b-v0.3-output-rotate-prefix}
    \end{subfigure}
    \caption{\textbf{Distribution of token-wise maximum values for $\mathbf{Q}$/$\mathbf{K}$/$\mathbf{V}$ in Mistral-7b-v0.3.}}
    \label{fig:mistral-7b-v0.3-output-all}
\end{figure}

We illustrate more token-wise maximum values distribution of other models. Details are as follows:
\begin{itemize}
    \item \textbf{Llama-2-7B}: Figure~\ref{fig:llama-2-7b-input-all} and Figure~\ref{fig:llama-2-7b-output-all} illustrate the distribution of input activation and $\mathbf{Q}$/$\mathbf{K}$/$\mathbf{V}$, respectively.
    \item \textbf{Llama-2-13B}: Figure~\ref{fig:llama-2-13b-input-all} and Figure~\ref{fig:llama-2-13b-output-all} illustrate the distribution of input activation and $\mathbf{Q}$/$\mathbf{K}$/$\mathbf{V}$, respectively.
    \item \textbf{Llama-3-8B}: Figure~\ref{fig:llama-3-8b-input-all} and Figure~\ref{fig:llama-3-8b-output-all} illustrate the distribution of input activation and $\mathbf{Q}$/$\mathbf{K}$/$\mathbf{V}$, respectively.
    \item \textbf{Llama-3-70B}: Figure~\ref{fig:llama-3-70b-input-all} and Figure~\ref{fig:llama-3-70b-output-all} illustrate the distribution of input activation and $\mathbf{Q}$/$\mathbf{K}$/$\mathbf{V}$, respectively.
    \item \textbf{Qwen-2-7B}: Figure~\ref{fig:qwen-2-7b-input-all} and Figure~\ref{fig:qwen-2-7b-output-all} illustrate the distribution of input activation and $\mathbf{Q}$/$\mathbf{K}$/$\mathbf{V}$, respectively.
    \item \textbf{Mistral-7B-v0.3}: Figure~\ref{fig:mistral-7b-v0.3-input-all} and Figure~\ref{fig:mistral-7b-v0.3-output-all} illustrate the distribution of input activation and $\mathbf{Q}$/$\mathbf{K}$/$\mathbf{V}$, respectively.
    
\end{itemize}

\newpage


\end{document}